\documentclass{article} 
\usepackage{iclr2027_conference,times}

\usepackage{amsmath,amsfonts,bm}

\def\eqref#1{equation~\ref{#1}}

\def\1{\bm{1}}

\DeclareMathAlphabet{\mathsfit}{\encodingdefault}{\sfdefault}{m}{sl}
\SetMathAlphabet{\mathsfit}{bold}{\encodingdefault}{\sfdefault}{bx}{n}

\usepackage{hyperref}
\usepackage{url}
\usepackage{array}
\usepackage[table]{xcolor}
\usepackage{booktabs,array,graphicx}
\usepackage{longtable}
\usepackage{caption}  
\usepackage{listings}  
\lstdefinestyle{mystyle}{
 backgroundcolor=\color{gray!10},
 basicstyle=\ttfamily\small,
 frame=single, framesep=6pt, framerule=0.5pt, rulecolor=\color{black!30},
 breaklines=true, breakatwhitespace=true, breakindent=0pt,
 showstringspaces=false, columns=flexible, captionpos=b,
 numberstyle=\tiny\color{gray},
 xleftmargin=10pt, xrightmargin=10pt, tabsize=4, extendedchars=true
}
\usepackage{enumitem} 
\usepackage{makecell} 

\title{How Do LLMs Change Predictions \\Under Negation?}

\author{%
  Jongwook Yoon$^{*}$ \quad
  Jongwon Lim$^{*}$ \quad
  Sungjib Lim \\
  \textbf{Woojin Cho$^{\ddagger}$ \quad Yohan Jo$^{\dagger}$} \\[4pt]
  Graduate School of Data Science, Seoul National University \\
  \texttt{\{jonguki,elijah0430,saint7451,f5foehn,yohan.jo\}@snu.ac.kr} 
}

\iclrfinalcopy 
\begin{document}
\addtocontents{toc}{\protect\setcounter{tocdepth}{-1}}

\maketitle
\lhead{Preprint.}

\begin{abstract}
Negation is an essential feature of human language, yet large language
models (LLMs) remain unreliable in processing it. We evaluate recent
open-source and closed-source LLMs on our negation benchmark and find
that, in 37--71\% of cases, they repeat the same answer under negation (e.g., ``Madrid'' for
``What is \textit{not} the capital of Spain?'').
To understand and address this brittleness, we mechanistically examine
how models operate under negation. Our main finding is that specialized
attention heads and MLP neurons jointly implement negation by
(1) suppressing retrieval of the original answer (e.g., ``Madrid'') while
(2) promoting a favored candidate within the answer category
(e.g., ``Paris'').
This contrasts with accounts of human negation processing, in which
information about the original answer helps to determine
what should be excluded.
Furthermore, we find that this difference from human processing is a key
source of negation failures: the model's mechanism relies on suppressing the
original answer rather than using it to determine what to exclude, 
so the model can repeat the original answer when suppression is too weak
or when a bias toward particular answers prevents it from selecting an
alternative.
To address this weakness in the model's negation mechanism, we propose a training objective that requires larger shifts in answer
preference for more confident original predictions, and show that it reduces negation failures with less degradation
of general capabilities than standard fine-tuning baselines.
Together, our results demonstrate how
mechanistic analysis can reveal why a linguistic capability fails and
guide training that targets the underlying limitation.
\footnote{We will release the dataset and code upon publication.}
\end{abstract}

\def\thefootnote{\fnsymbol{footnote}}
\footnotetext[1]{Equal contribution.}
\footnotetext[2]{Corresponding author.}
\footnotetext[3]{Visiting intern from Department of Computer Science and Engineering, Seoul National University.}


\section{Introduction}

Negation is an essential feature of human language.
However, large
language models (LLMs) continue to struggle with negation
\citep{truong2023naysayers,garciaferrero2023notdataset,ravichander-etal-2022-condaqa}, and our evaluation
shows that these difficulties persist even in recent production-class
models (\S~\ref{sec:evaluation}).
The challenge is that a negated prompt such as ``What is \textit{not} the capital of Spain?'' requires not only
avoiding the original answer ``Madrid'', but also selecting an
appropriate alternative, such as another city (e.g., ``Paris'').
In other words, processing negation requires (1) using the knowledge
underlying the original answer to select an alternative within the
same answer category (e.g., a city name), (2) while still rejecting the original answer itself.
Indeed, research on human negation processing suggests that humans process
negation by first retrieving the original meaning and then using it to
determine what should be excluded~\citep{fischler1983brain,mayo2004successful,nieuwland2008truth},
suggesting that the knowledge of the original answer can inform, rather than hinder,
negation processing.
However, whether LLMs rely on a similar mechanism remains unclear.

In this paper, we provide a mechanistic explanation of how LLMs internally
process negation and use this understanding to diagnose and address their
failures. We begin with a behavioral examination of recent open-source
and closed-source models using negation tests drawn from six benchmarks
covering factual QA, visual QA, and logical reasoning
(\S~\ref{sec:evaluation}).
Our evaluation reveals a recurring failure mode: in 37.4--71.1\% of originally
correct cases, models repeat the same answer under negation.
Crucially, this failure becomes more frequent when the model is more
confident in its original answer, contrary to the expectation that knowing
the answer should help identify what to exclude.
Furthermore, models often favor a particular answer under negation. When this is also the original answer, they are especially
likely to repeat it.
Together, these findings suggest that models may not reliably use
knowledge of the original answer as a basis for exclusion.

\begin{figure}[t]
    \centering
    \includegraphics[width=\linewidth]
    {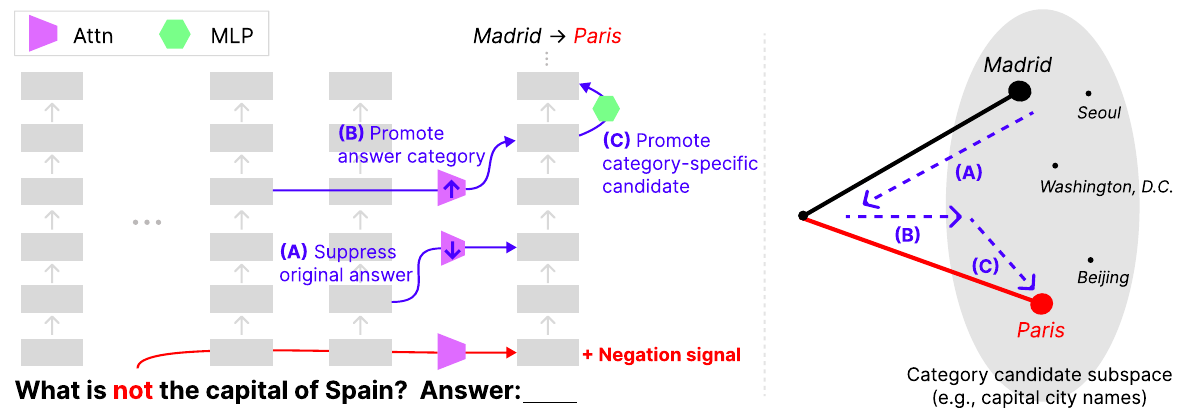}
\caption{
Overview of LLMs' negation mechanism.
\textbf{Left:} The negation signal triggers three downstream changes:
\textbf{(A)} suppressing the original answer,
\textbf{(B)} promoting answer category information, and
\textbf{(C)} promoting a category-specific alternative.
\textbf{Right:} These changes preserve the answer category subspace
while shifting preference to an
alternative answer (``Paris'').}    \label{fig:mechanism}
\end{figure}

To understand the mechanism underlying these failures, we proceed in three steps. First, to identify where
negation begins to affect the model's answer prediction, we trace
the information flow from the negation token to the prediction token position
using activation
patching~\citep{meng2022locating,actpatch} (\S~\ref{sec:negation-localization}).
We find that negation reaches the final prediction position to shift the representation in a similar direction across examples:
subtracting this direction from a negated prompt's representation shifts
downstream computation and the prediction back toward those of the original prompt.
We refer to this causal direction as the \emph{negation signal}.

We then analyze how downstream computation changes to generate an alternative answer (\S~\ref{sec:selective-correction}). Interestingly, we discover that instead of using knowledge of the original answer to determine what should be excluded, as humans do, LLMs perform negation by merely suppressing the original answer and promoting a category-specific alternative. Specifically, we identify three distinct yet complementary mechanisms as in Figure~\ref{fig:mechanism}: \textbf{(A)} weakening attention heads that retrieve the original answer (e.g., ``Madrid''), \textbf{(B)} strengthening attention heads that retrieve broad answer category information (e.g., the answer should be a city name), and \textbf{(C)} activating MLP neurons that promote particular candidates within that category (e.g., ``Paris'' over other capitals). To examine how these changes jointly shape the model's representations, we use Distributed Alignment Search (DAS)~\citep{DAS}.
We find that negation largely preserves answer category information while selectively modifying representations within the candidate-preference subspace.

Lastly, to diagnose negation failures, we compare successful cases with failure cases in which the model repeats the original answer under negation (\S~\ref{sec:failure-diagnosis}).
We find that because the mechanism relies on suppressing the original answer, rather than using the original answer to determine what should be excluded, it becomes vulnerable when suppression is insufficient relative to the support for the original answer. This support correlates with the model's confidence in that answer.
Based on this diagnosis, we propose a training objective that requires stronger corrections for more confident original answers. Compared with multiple baselines, our objective more effectively reduces negation failures while preserving performance on non-negated prompts.

Together, we demonstrate a distinct mechanism for negation. Unlike human
accounts in which the original meaning informs what to exclude, LLMs
suppress the original prediction while promoting a favored alternative
within the same category. This explains their failures: the correction
can be too weak to overturn a strongly supported original prediction.
Guided by this diagnosis, we develop a confidence-adaptive training
objective that reduces negation failures.

\section{Evaluating Negation Processing in LLMs}
\label{sec:evaluation}

To assess how well current LLMs process negation, we systematically evaluate their ability to generate appropriate alternatives under negation across multiple domains.
We find that both open-source and production-class closed-source models struggle with negation.
While these models tend to generate appropriate alternatives rather than unrelated words, they often fail to change their answers under negation, instead repeating the answers they gave to the corresponding non-negated prompts.

\paragraph{Dataset construction.}
We construct paired prompts that isolate the effect of negation while
preserving the remaining context.
Our dataset draws on six benchmarks, with two representing each domain:
PopQA~\citep{popqa} and RippleEdits~\citep{rippleedit} for factual
association, PhantomWiki~\citep{phantomwiki} and
SynthWorlds~\citep{synthworld} for logical reasoning, and
GQA~\citep{GQA} and PTR~\citep{PTR} for visual question answering.
To cover diverse surface forms of negation, we construct four negated
counterparts for each example by inserting \textit{not}, \textit{n't},
\textit{never}, or \textit{by no means}.
To assess whether models generate alternatives of the appropriate type,
we use source metadata to identify the type of answer each question
requests (e.g., a person name, a sport, or a letter), which we call its
\emph{answer category}.

\begin{table}[t]
\centering
\begingroup
\small\mdseries\color{black}
\setlength{\tabcolsep}{3pt}
\renewcommand{\arraystretch}{1.18}

\newcommand{\resultpair}[2]{%
  \makebox[2.4em][r]{#1}%
  \hspace{0.2em}/\hspace{0.2em}%
  \makebox[2.4em][l]{#2}%
}
\newcommand{\pairheading}{%
  \shortstack{\scriptsize Ans. change /\\\scriptsize Cat. pres.}%
}
\def\avgseparator{%
  \hspace{5pt}{\color{black!40}\vrule width 0.35pt}\hspace{5pt}%
}

\caption{Answer change and category preservation rates on the test
split of our benchmark (\%). Averages are computed over six benchmarks for multimodal
models and four for text-only models.}
\label{tab:model_stats}

\par\vspace{6pt}

\resizebox{\linewidth}{!}{%
\begin{tabular}{@{}l@{\hspace{8pt}}
*{6}{c}
@{\avgseparator}c
@{\hspace{7pt}}>{\columncolor{black!8}}r@{}}
\toprule
& \multicolumn{2}{c}{Factual Association}
& \multicolumn{2}{c}{Logical Reasoning}
& \multicolumn{2}{c@{\avgseparator}}{Visual Question Answering}
& \multicolumn{2}{c}{Average} \\
\cmidrule(lr){2-3}
\cmidrule(lr){4-5}
\cmidrule(lr){6-7}
\cmidrule(lr){8-9}
& PopQA
& RippleEdits
& PhantomWiki
& SynthWorlds
& GQA
& PTR
& Negated
& \multicolumn{1}{c@{}}{} \\
\cmidrule(lr){2-2}
\cmidrule(lr){3-3}
\cmidrule(lr){4-4}
\cmidrule(lr){5-5}
\cmidrule(lr){6-6}
\cmidrule(lr){7-7}
\cmidrule(lr){8-8}
Model
& \pairheading
& \pairheading
& \pairheading
& \pairheading
& \pairheading
& \pairheading
& \pairheading
& \multicolumn{1}{c@{}}{%
    \shortstack{\scriptsize Original\\\scriptsize acc.}%
  } \\
\midrule
\multicolumn{9}{@{}l}{\textit{Open-source models}} \\
\addlinespace[2pt]
Gemma 3-4B-IT
& \resultpair{40.8}{98.5}
& \resultpair{14.1}{95.7}
& \resultpair{66.4}{90.8}
& \resultpair{35.0}{90.3}
& \resultpair{46.3}{93.0}
& \resultpair{31.8}{97.3}
& \resultpair{39.1}{94.3}
& 32.6 \\
Gemma 3-12B-IT
& \resultpair{58.1}{99.7}
& \resultpair{31.1}{95.9}
& \resultpair{50.4}{96.2}
& \resultpair{42.3}{95.2}
& \resultpair{58.3}{94.9}
& \resultpair{44.4}{99.1}
& \resultpair{47.4}{96.8}
& 41.9 \\
Gemma 3-27B-IT
& \resultpair{68.2}{99.9}
& \resultpair{44.8}{98.8}
& \resultpair{70.1}{99.6}
& \resultpair{42.9}{95.4}
& \resultpair{77.5}{94.1}
& \resultpair{72.5}{98.9}
& \resultpair{62.6}{97.8}
& 44.8 \\
Qwen3.5-9B
& \resultpair{55.4}{99.3}
& \resultpair{52.5}{89.9}
& \resultpair{63.3}{99.6}
& \resultpair{24.5}{95.5}
& \resultpair{55.2}{96.1}
& \resultpair{66.2}{98.9}
& \resultpair{52.9}{96.6}
& 47.2 \\
Llama 3.1-8B-Instruct
& \resultpair{43.3}{83.4}
& \resultpair{13.5}{82.9}
& \resultpair{26.8}{97.8}
& \resultpair{32.2}{84.5}
& \multicolumn{2}{c@{\avgseparator}}{\textit{Text only}}
& \resultpair{28.9}{87.2}
& 39.3 \\
OLMo 3-7B-Instruct
& \resultpair{56.2}{89.7}
& \resultpair{32.0}{93.4}
& \resultpair{65.3}{92.8}
& \resultpair{30.5}{97.6}
& \multicolumn{2}{c@{\avgseparator}}{\textit{Text only}}
& \resultpair{46.0}{93.4}
& 32.4 \\
\midrule
\multicolumn{9}{@{}l}{\textit{Closed-source models}} \\
\addlinespace[2pt]
GPT-5.6 Luna
& \resultpair{55.3}{92.3}
& \resultpair{43.6}{89.5}
& \resultpair{39.1}{95.4}
& \resultpair{31.6}{100.0}
& \resultpair{61.4}{97.5}
& \resultpair{57.1}{98.4}
& \resultpair{48.0}{95.5}
& 52.3 \\
Claude Sonnet 5
& \resultpair{66.3}{98.4}
& \resultpair{51.0}{97.0}
& \resultpair{67.0}{87.3}
& \resultpair{34.9}{92.1}
& \resultpair{73.7}{97.9}
& \resultpair{68.6}{99.4}
& \resultpair{60.3}{95.3}
& 54.3 \\
\bottomrule
\end{tabular}%
}
\endgroup
\end{table}

To ensure the quality of our benchmark, we evaluate each negated question based on three criteria: whether it is grammatically well-formed, semantically unambiguous, and preserves the meaning of the original question under negation.
The resulting dataset contains 105,048 examples, which we split into
training, validation, and test sets at a 4:1:5 ratio.
We use the training and validation splits for mechanistic analyses and
reserve the test split for the final evaluation of our training method.
Construction rules and validation details are provided in
Appendix~\ref{app:benchmark}.

\paragraph{Models.}
We evaluate both open-source and production-class closed-source models.
The six open-source models are Gemma 3-\{4B,12B,27B\}-IT,
Qwen3.5-9B, Llama 3.1-8B-Instruct, and OLMo 3-7B-Instruct, and the closed-source models are GPT-5.6 Luna and Claude Sonnet 5.

\paragraph{Evaluation.}
For each pair, we query the model separately with the original and negated
prompts, keeping all other input unchanged.
We evaluate the three metrics sequentially, with each subsequent metric assessed only on cases that pass the preceding metric.
First, we assess \emph{accuracy on the original questions} against the
source benchmarks' reference answers.
Second, \emph{answer change rate} measures the fraction of originally correct
cases in which the answer changes semantically under negation.
Third, \emph{category preservation rate} measures the fraction of these
changed answers that belong to the same answer category as the original answer.
Further details are provided in Appendix~\ref{app:evaluation_details}.

\paragraph{Results.}
Table~\ref{tab:model_stats} shows that LLMs can select
alternatives that are consistent with the answer category, rather than simply
produce unrelated answers. When models change their answers,
category preservation rates are high across models (87.2--97.8\%).
Nevertheless, models often repeat the original answer under negation, with answer change rates of only 28.9--62.6\%.
This occurs even in production-class closed-source models including GPT-5.6 Luna and Claude Sonnet 5.

\begin{figure}[t]
  \centering
  \includegraphics[width=0.9\linewidth]{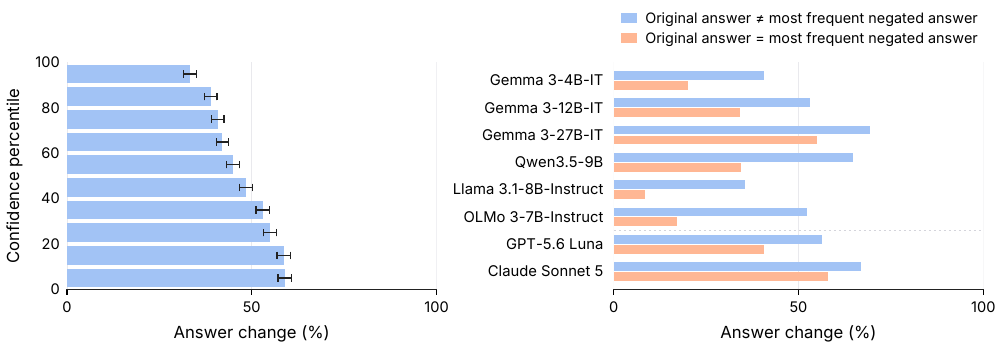}
\caption{
\textbf{Left:} Answer change rates decrease with confidence
in the original answer (mean token log-probability; results averaged across
six open-source models).
\textbf{Right:} Answer change rates are also lower when the original answer is the model's
most common response under negation.
}
  \label{fig:answer-bias-and-confidence}
\end{figure}

We further find that higher original answer confidence is associated
with more frequent repetition
(Figure~\ref{fig:answer-bias-and-confidence}, left).
We also observe that models tend to generate a particular answer frequently under negation.
We therefore compare cases where this most frequent negated answer matches the original answer with those where it does not (Figure~\ref{fig:answer-bias-and-confidence}, right).
The answer change rate drops substantially when the two match.
This suggests that such a preference often facilitates successful negation by providing an alternative, but can substantially hinder negation when the preferred answer is itself the original answer that must be rejected.
We refer to this systematic preference as \emph{negation bias}.
Additional results are provided in Appendix~\ref{app:additional-results-evaluation}.

Together, these findings suggest that LLMs possess a mechanism for
selecting appropriate alternatives under negation, but that this
mechanism remains brittle. We next investigate how this mechanism generates
alternatives and why it sometimes fails to override the original answer.

\section{Localizing the Effect of Negation}
\label{sec:negation-localization}

To understand how negation changes answer generation, we first localize
\textit{where} it affects the model's computation. We trace how its effect
is transmitted 
to the final prompt token
(\S~\ref{sec:information_flow}), where it shifts middle-layer
representations in a similar direction across examples.
Removing this direction from the final prompt token causally reverses
subsequent attention and MLP changes
(\S~\ref{sec:negation_signal}), allowing us to focus the following
analyses on these components.

\paragraph{Setup and notation.}
We conduct our analysis on four of the open-source models evaluated in
\S~\ref{sec:evaluation}: Gemma 3-\{4B, 12B\}-IT,
Qwen3.5-9B, and Llama 3.1-8B-Instruct.
We use Gemma 3-12B-IT for our main analyses and report results for the
remaining models in Appendix~\ref{app:additional-results}.
Here, we analyze only cases where models generate an appropriate
alternative under negation.
To understand how negation interacts with the computation that generates
the answer, we distinguish four types of prompt tokens:
(1) \textit{answer tokens}, such as ``Spain'' in ``What is the capital
of Spain?'', which contain information about the original answer;
(2) \textit{answer category tokens}, such as ``the capital of'', which
specify the answer category;
(3) \textit{negation tokens} in the negated prompt; and
(4) \textit{the final prompt token}, where the prediction is formed.

Each example $i$ pairs an original prompt $x_i$ with a negated prompt
$x_{i,f}^{\mathrm{neg}}$, where $f$ denotes the negation form.
We align corresponding token positions across the two prompts while
ignoring the inserted negation expression, and define the negation-induced
change as
$\Delta\mathbf h_{i,f}^{\ell,t}
=\mathbf h_{\ell,t}(x_{i,f}^{\mathrm{neg}})
-\mathbf h_{\ell,t}(x_i)$,
where $\mathbf h_{\ell,t}(x)$ is the residual-stream representation at
layer $\ell$ and aligned position $t$.

For causal interventions, we report two metrics:
(1) \emph{original-answer match rate}, the fraction of cases where the
intervened model outputs the same answer as the original prompt, and
(2) \emph{normalized log-probability gap shift}, which measures how far
the intervention shifts the log-probability gap between the original
and negated answers back toward that under the original prompt.

Let $a_i$ and $b_{i,f}$ denote the original and negated answers.
We compute each answer's full-sequence log-probability,
$s(a\mid x)=\sum_{k=1}^{|a|}\log P(a_k\mid x,a_{<k})$,
and define the log-probability gap as
$g_{i,f}(x)=s(b_{i,f}\mid x)-s(a_i\mid x)$.
Let $g_{i,f}^{\mathrm{int}}$ denote this gap after intervention on the
negated prompt. The normalized log-probability gap shift is
\[
\frac{1}{|\mathcal D|}
\sum_{(i,f)\in\mathcal D}
\frac{
g_{i,f}(x_{i,f}^{\mathrm{neg}})-g_{i,f}^{\mathrm{int}}
}{
g_{i,f}(x_{i,f}^{\mathrm{neg}})-g_{i,f}(x_i)
},
\]
where $\mathcal D$ denotes the set of evaluated pairs.


\subsection{Tracing the Information Flow of Negation}
\label{sec:information_flow}

\paragraph{Method.}
To trace how information from the negation tokens reaches the final
prompt token, we use activation patching~\citep{actpatch}.
We examine attention paths carrying information
(1) from the negation tokens to other token groups and
(2) from each token group to the final prompt token, with one overlap
(e.g., a direct negation-to-final-token path).
For each path, we replace only the corresponding part of the attention
output with that from the paired prompt, leaving the remaining output
unchanged. We perform this intervention in both directions and measure
normalized log-probability gap recovery.
For each path, we apply the intervention within a four-layer sliding window with a stride of two layers.
Detailed procedures and settings are in
Appendix~\ref{app:information_flow}.

\begin{figure*}[t]\centering
\includegraphics[width=\textwidth]{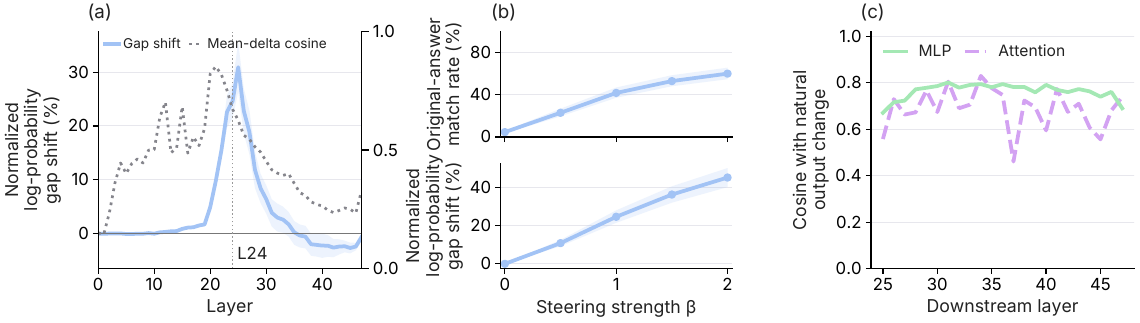}
\caption{Removing the negation direction in Gemma 3-12B-IT.
\textbf{(a)} Layerwise effects of steering with the negation signal ($\beta=1$), overlaid with negation-delta similarity.
\textbf{(b)} Causal effects of steering with the negation signal at (L24).
\textbf{(c)} Cosine similarity between attention and MLP output changes induced by
natural negation and by steering ($\beta=3$).
Effects are averaged over six benchmarks. Shading indicates 95\% bootstrap intervals.}\label{fig:negation-signal}
\end{figure*}

\paragraph{Results.}
The results in Figure~\ref{fig:gemma4-information-flow} suggest that the
effect of negation reaches the final prompt token mainly through other
prompt tokens rather than directly. Notably, the direct path from the
negation tokens to the final prompt token has relatively little effect (1.3\% increase).
Among paths originating from the negation tokens, those targeting
answer category tokens (e.g., ``the capital of'') show a large
effect peaking at early layers (Figure~\ref{fig:gemma4-information-flow} left, L8--11), followed by effects from answer category and other tokens to the
final prompt token in the middle layers (Figure~\ref{fig:gemma4-information-flow} right, L22--25). This pattern  suggests that
negation is first transmitted through answer category tokens before affecting the
representation at the final prompt token.
Answer-token paths also show large effects (Figure~\ref{fig:gemma4-information-flow}, green line), but they peak later and reflect
a different process. Separate interventions on attention weights and
answer-token value vectors show that these effects arise mainly from
changes in attention weight attending to answer tokens, rather than changes in the
information carried by their value vectors
(Appendix~\ref{app:information_flow}).

\subsection{A Causal Negation Signal at the Final Prompt Token}
\label{sec:negation_signal}

To identify where alternatives are generated, we ask whether the negation
information reaching the final prompt token already specifies an alternative
answer or instead provides a signal for generating one in subsequent
computation. To investigate this, we compare the directions of
negation-induced hidden-state changes across samples with different answers.
We find that these changes are highly aligned: the average cosine similarity to the mean negation-induced change reaches $0.853$ at L22 across the six benchmarks, within the same middle-layer range where activation patching shows negation affecting this token (Figure~\ref{fig:negation-signal}(a)).
This suggests a consistent negation-related component across samples, motivating us to test whether removing this component reverses the subsequent effects of negation.

We test this by subtracting the mean negation-induced difference from
the negated prompt's representation at each layer. The effect of removal
increases in the middle layers and peaks around L25.
At this layer, removing the direction restores the original answer in up to 48.6\% of cases (Figure~\ref{fig:negation-signal}(b)).
Importantly, changing only the representation at the final prompt token
also reverses downstream attention and MLP output changes toward those
of the original prompt (Figure~\ref{fig:negation-signal}(c)).
We therefore refer to this direction as the \emph{negation signal}.
Detailed procedures and additional results are provided in
Appendix~\ref{app:negation_signal}.

\section{How LLMs Generate Alternatives under Negation}
\label{sec:selective-correction}

We next investigate the specific mechanisms by which the negation signal changes downstream computation
to produce an alternative in a related answer category.
We identify specific attention heads and MLP neurons that contribute
to suppressing the original answer and promoting alternatives
(\S~\ref{sec_5_characterization}). We then show how these changes
are reflected in the model's representations: answer category
information is largely preserved, while preference among answers
changes (\S~\ref{sec_5_representation}).

\subsection{Components that Implement Negation}
\label{sec_5_characterization}

\paragraph{Method.}
To understand how the negation signal changes answer generation, we
identify downstream attention heads and MLP neurons involved in this
change at the final prompt token.
For attention heads, we rank heads by the magnitude of their
negation-induced attention change. For each paired prompt, we score
head $h$ as
$S_h^{\mathrm{attn}}=\|\Delta\boldsymbol{\alpha}_h\|$,
where $\Delta\boldsymbol{\alpha}_h$ is the change in its attention
weights from the final prompt token.
For MLP neurons, we instead identify activations contributing to the prediction of the alternative answer using attribution patching
~\citep{syed-etal-2024-attribution}. We score neuron $j$ as
$S_j^{\mathrm{MLP}}=\Delta m_j\,(\partial g/\partial m_j)$,
where $\Delta m_j$ is its activation change and $g$ measures the
log-probability gap between the alternative and original answers.
We average these scores over training examples to rank components.
Detailed procedures are provided in
Appendix~\ref{app:component_identification}.

\begin{figure*}[t]
  \centering
  \includegraphics[width=\textwidth]{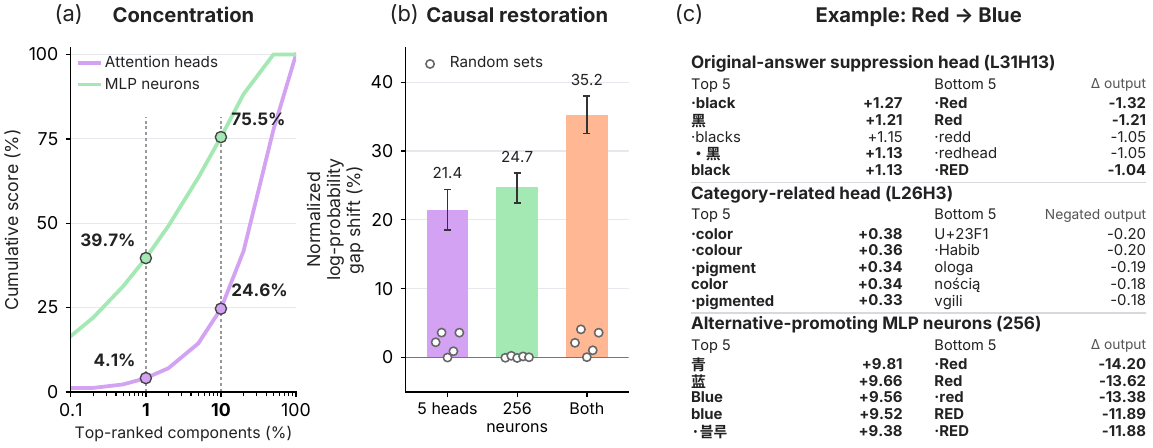}
  \caption{Components involved in forming alternatives in Gemma 3-12B-IT. (a) Cumulative head and neuron scores. (b) Effects of restoring their non-negated values, averaged across six benchmarks; dots show matched random controls. (c) Unfiltered logit lens top/bottom five for an illustrative example.}
  \label{fig:mechanism_analysis}
\end{figure*}

\paragraph{Results.}
Averaging these scores over training examples reveals a clear pattern:
a small set of attention heads and MLP neurons consistently ranks highly
across datasets and answer categories. The top 1\% of downstream neurons
(3,533 neurons) accounts for 39.7\% of the positive attribution score
(Figure~\ref{fig:mechanism_analysis} (a)).
Restoring selected head outputs or neuron activations to their non-negated
values in held-out prompts produces larger changes in answer preference
than matched random controls (Figure~\ref{fig:mechanism_analysis} (b)).

To characterize their roles, we examine attention to answer category tokens and use logit lens~\citep{nostalgebraist2020logitlens} to interpret component outputs.
These analyses suggest three distinct yet complementary mechanisms,
which we test through causal interventions.

First, several middle-layer heads (e.g., L31H13) reduce attention to
answer tokens under negation. Their non-negated outputs support the
original answer across different answer categories, but this support
decreases under negation (Figure~\ref{fig:mechanism_analysis} (c)).
Following prior work on factual retrieval
\citep{geva-etal-2023-dissecting}, we interpret this as weakened retrieval
of the original answer. Restoring the attention outputs from answer
tokens to their non-negated values increases preference for that answer,
supporting this interpretation
(Appendix~\ref{app:additional-results-components_1}).

Second, other heads increase attention to answer category tokens.
Their logit lens outputs support multiple candidates within the relevant
category without consistently favoring a particular alternative
(Figure~\ref{fig:mechanism_analysis}(c)).
Removing these outputs reduces preference for answers across the category,
rather than selectively for the model-generated alternative
(Appendix~\ref{app:additional-results-components_2}).
This suggests that these heads promote the answer category rather than
choose a particular replacement.

Third, high-attribution MLP neurons promote particular alternatives within
the relevant category. Restoring their non-negated activations weakens
preference for the model-generated alternative over other candidates
in that category, whereas amplifying their negation-induced changes
strengthens it.
Importantly, their output changes tend to favor similar candidates
within each category even when the original answer differs.
This suggests that these neurons promote preferred candidates rather
than consistently adjust their choice to exclude the original answer
(Appendix~\ref{app:additional-results-components_3}).

\begin{figure}[t]
    \centering
    \includegraphics[width=0.9\linewidth]{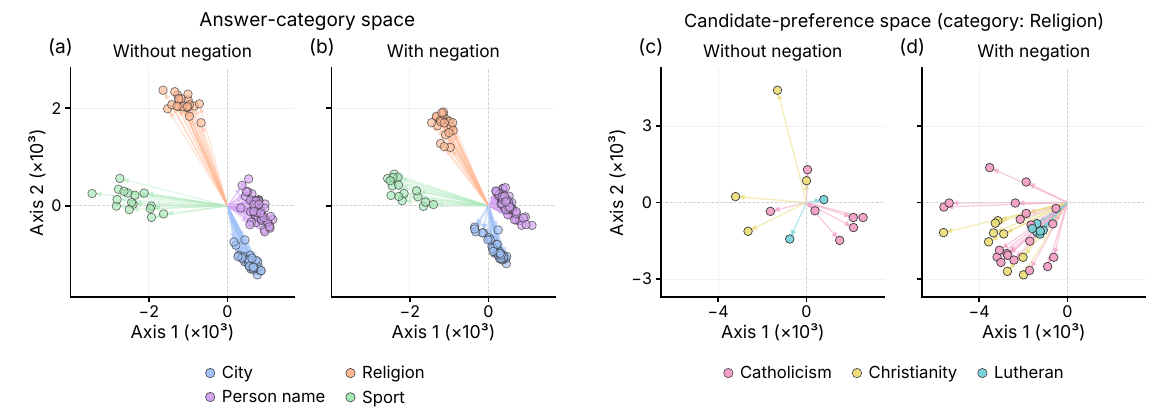}
    \caption{Negation largely preserves the shared answer category structure
    (a--b), while redirecting candidate-space representations toward a specific direction (c--d).
    Arrows show independently computed activations without centering or
    normalization.}
    \label{fig:selective-correction-geometry}
\end{figure}

Together, these results suggest that the model forms alternatives by
weakening retrieval of the original answer while promoting answer category
information and preferred candidates within that category.
This can produce an appropriate alternative, but may fail when the
original answer is itself among the favored candidates, as shown in Figure~\ref{fig:answer-bias-and-confidence}, right.

\subsection{How Negation Reshapes the Representation}
\label{sec_5_representation}

\paragraph{Method.}
To understand how these mechanisms jointly change answer generation,
we examine what information they preserve and modify in the representation
at the final prompt token. We hypothesize that (1) answer category information
is largely preserved while (2) preference among candidates changes, as
illustrated in Figure~\ref{fig:mechanism}, right.

To distinguish these two types of information, we build on Distributed
Alignment Search (DAS; \citealp{DAS}), which uses gradient-based optimization
to find subspaces where interventions produce a specified change in the
model's prediction.
Using separate objectives, we learn an answer category space $C$, where
interventions increase preference for answers from another category,
and a candidate-preference space $L_s$, where interventions increase
preference for a different answer within category $s$.
We learn both spaces from prompts without negation and keep them fixed
when comparing the magnitude and direction of negation-induced changes
within each space.
The optimization objectives and intervention details are provided in
Appendix~\ref{app:subspace_experiment}.



\paragraph{Results.}
Comparing matched original and negated prompts, we find that negation
largely preserves answer category information in $C$ while changing
candidate preference in $L_s$
(Figure~\ref{fig:selective-correction-geometry}).
This pattern is consistent with the joint effects of the mechanisms
identified above: answer category heads offset the effects of reduced
answer retrieval on category information, while original-answer
suppression and alternative-promoting neurons jointly shift candidate
preference.

Interestingly, negation-induced changes in $L_s$ tend to align in a
similar direction within each answer category, even when the original
answers differ. To identify which answers this direction favors, we
intervene along it and measure changes in answer scores.
We find that it favors the answer most frequently generated under
negation (i.e., \textit{negation bias}) within that category
(Appendix~\ref{app:subspace_experiment}).
Together with the neuron analysis, this suggests that the model retains
the category information needed to generate alternatives, but does not
reliably adjust its candidate preferences to exclude the original answer.


\section{Why Negation Fails and How to Improve It}
\label{sec:negation-failure-and-solution}

We next analyze failure cases, and find that the mechanism remains active but does not sufficiently suppress 
the original answer (\S~\ref{sec:failure-diagnosis}). Based on this diagnosis, we introduce a
training objective that requires stronger corrections when the
original prediction is more confident (\S~\ref{sec:negation-training}).

\subsection{Diagnosing Failure Cases}
\label{sec:failure-diagnosis}

We ask whether negation failures occur because the mechanisms
identified above are absent, or because their effects are too weak to
change the prediction. Comparing successful and repeated-answer cases,
we find that negation-induced changes remain present in failures, but
are weaker (Appendix~\ref{app:dianosis}). For example, L31H13's attention to answer tokens decreases
from 36.3\% to 8.6\% in successful cases, but only from 38.1\% to
17.2\% in failures. 

We then test whether strengthening the identified changes can alter
failed predictions. Removing the selected heads' answer-token
contributions lowers the original-answer match rate from 100\% to
94.9\%, while doubling the selected neurons' negation-induced changes
lowers it to 58.6\%. Combining both yields 57.6\%
(Appendix~\ref{app:dianosis}).
Thus, the mechanism can remain active in failure cases but be
insufficient to overturn the original prediction.
Together with the association between original-answer confidence and
repetition (\S~\ref{sec:evaluation}), this motivates training that
requires stronger corrections for more confident original predictions.

\subsection{Improving Negation through Confidence-Adaptive Training}
\label{sec:negation-training}

Our analysis suggests that negation can fail even when correction is active,
because it is too weak to overcome support for the original answer.
We therefore propose \emph{Adaptive Logit Inversion Training} (ALiT),
which requires stronger correction for more confident original predictions.
Rather than specifying a particular alternative, we extrapolate the
model's negation-induced logit change until an alternative is preferred
over the original answer by a confidence-dependent margin, then train
the model to match the resulting distribution.
\paragraph{Constructing the training target.}
Using a frozen base model, we compute logits at the first answer position
and extrapolate their change under negation:
\begin{equation}
\widetilde{\textbf{logit}}_i(\beta)
=
\textbf{logit}_0(x_i)
+
\beta\left[
\textbf{logit}_0(x_{i, f}^{\mathrm{neg}})
-
\textbf{logit}_0(x_i)
\right],
\qquad \beta\geq 1.
\label{eq:alit-target}
\end{equation}
At $\beta=1$, the target equals the unmodified negated logits; larger
values amplify the model's existing change.
Let $m_i$ be the gap between the two highest logits on the original
prompt. We choose the smallest $\beta\geq1$ for which another token
exceeds the original top token by at least $\alpha m_i$, where $\alpha$
controls the required margin. Thus, more confident original predictions
require a larger margin in favor of an alternative.
The coefficient cap and handling of infeasible margins are described
in Appendix~\ref{app:training}.

\paragraph{Training objective.}
Let $\widetilde p_i$ be the softmax distribution of the resulting target
logits. We train the model to match this target while penalizing changes
to its original predictions:
\begin{equation}
\mathcal L
=
\mathbb E_i\left[
D_{\mathrm{KL}}\!\left(
p_\theta(\cdot\mid x_i^{\mathrm{neg}})
\,\Vert\,\widetilde p_i
\right)
\right]
+
\lambda\mathcal L_{\mathrm{orig}},
\label{eq:confidence-adaptive-loss}
\end{equation}
where $\mathcal L_{\mathrm{orig}}$ penalizes deviations from the frozen
base model on non-negated prompts. Targets and confidence margins are
fixed before training. Full loss definitions and optimization settings
are provided in Appendix~\ref{app:training}.

\paragraph{Setup.}
We post-train Gemma 3-12B-IT on the PopQA subset of our benchmark, treating it as in-domain and other benchmarks as out-of-domain. We compare ALiT with other training algorithms such as supervised fine-tuning
(SFT; \citealp{she-etal-2023-scone}), unlikelihood training
(UL; \citealp{hosseini-etal-2021-understanding}), and Direct
Preference Optimization (DPO; \citealp{rafailov2023direct}),
alongside the unmodified base model.
Training hyperparameters and baseline target construction are
detailed in Appendix~\ref{app:training}.

For evaluation, We use the held-out test split using the three metrics from
\S~\ref{sec:evaluation}: answer change, category preservation, and
original-question accuracy. We treat PopQA as in-domain (ID) and the
remaining five benchmarks as out-of-domain (OOD). We additionally measure \emph{answer diversity} within each category
group as one minus the fraction of predictions accounted for by the most
frequent answer, and average this score across groups in proportion to
their size. To assess general capability, we evaluate MMLU-Pro, HellaSwag, GSM8K,
and IFEval, and report their average performance.

  \begin{table}[t]
  \centering
  \begingroup
  \footnotesize
  \setlength{\tabcolsep}{3.5pt}
  \renewcommand{\arraystretch}{1.13}

  \definecolor{positive}{RGB}{35,100,170}
  \definecolor{negative}{RGB}{185,55,55}
  \definecolor{neutral}{RGB}{110,110,110}

  \newcommand{\alitdiff}[3]{%
    \shortstack{#1\\{\scriptsize\textcolor{#3}{($#2$)}}}%
  }

  \caption{
Training results for Gemma 3-12B-IT (\%).
  ID denotes PopQA, and OOD denotes five other benchmarks.
  }
  \label{tab:negation_training}

  \resizebox{\linewidth}{!}{%
  \begin{tabular}{@{}l
      ccc >{\columncolor{black!8}}c
      @{\hspace{9pt}}
      ccc >{\columncolor{black!8}}c
      @{\hspace{9pt}} c@{}}
  \toprule
  & \multicolumn{4}{c}{In-domain (ID)}
  & \multicolumn{4}{c}{Out-of-domain (OOD)}
  & General Capability \\
  \cmidrule(lr){2-5}
  \cmidrule(lr){6-9}
  \cmidrule(lr){10-10}
  Method
  & \shortstack{Answer\\change}
  & \shortstack{Category\\preservation}
  & \shortstack{Answer\\diversity}
  & \shortstack{Original\\accuracy}
  & \shortstack{Answer\\change}
  & \shortstack{Category\\preservation}
  & \shortstack{Answer\\diversity}
  & \shortstack{Original\\accuracy}
  & Avg. \\
  \midrule

  Base
  & 58.14 & 99.71 & 90.90 & 21.90
  & 46.93 & 96.64 & 80.05 & 37.15
  & 78.04 \\

  \addlinespace[5pt]

  \textbf{ALiT (ours)}
  & 96.55 & 99.86 & 86.54 & 22.10
  & 84.96 & 97.63 & 76.26 & 37.40
  & 77.68 \\

  \midrule

  Unlikelihood
  & \alitdiff{99.95}{+3.40}{positive}
  & \alitdiff{53.97}{-45.89}{negative}
  & \alitdiff{75.21}{-11.33}{negative}
  & \alitdiff{21.16}{-0.94}{negative}
  & \alitdiff{89.41}{+4.45}{positive}
  & \alitdiff{57.68}{-39.95}{negative}
  & \alitdiff{84.46}{+8.20}{positive}
  & \alitdiff{37.40}{0.00}{neutral}
  & \alitdiff{77.95}{+0.27}{positive} \\
  \addlinespace[3pt]

  SFT
  & \alitdiff{96.84}{+0.29}{positive}
  & \alitdiff{100.00}{+0.14}{positive}
  & \alitdiff{8.03}{-78.51}{negative}
  & \alitdiff{21.64}{-0.46}{negative}
  & \alitdiff{86.52}{+1.56}{positive}
  & \alitdiff{99.06}{+1.43}{positive}
  & \alitdiff{39.11}{-37.15}{negative}
  & \alitdiff{36.64}{-0.76}{negative}
  & \alitdiff{76.98}{-0.70}{negative} \\
  \addlinespace[3pt]

  DPO
  & \alitdiff{100.00}{+3.45}{positive}
  & \alitdiff{42.76}{-57.10}{negative}
  & \alitdiff{73.61}{-12.93}{negative}
  & \alitdiff{21.97}{-0.13}{negative}
  & \alitdiff{99.23}{+14.27}{positive}
  & \alitdiff{52.63}{-45.00}{negative}
  & \alitdiff{83.71}{+7.45}{positive}
  & \alitdiff{37.16}{-0.24}{negative}
  & \alitdiff{77.36}{-0.32}{negative} \\

  \bottomrule
  \end{tabular}%
  }
  \endgroup
  \end{table}

\paragraph{Results.}
Table~\ref{tab:negation_training} shows that ALiT substantially improves
answer change while preserving answer category consistency and performance
on the original questions. Answer change increases from 58.1\% to 96.6\%
in-domain and from 46.9\% to 85.0\% out-of-domain, while category
preservation remains above 97\%. Original accuracy and general-capability
performance also remain close to those of the base model. The baselines, on the other hand, reveal different trade-offs. UL and DPO achieve high answer
change, but substantially reduce category preservation. SFT preserves the answer category and achieves similarly high answer change, but sharply
reduces answer diversity, suggesting that it learns to rely on a small set
of alternatives.

Figure~\ref{fig:training-answer-bias} makes this difference more explicit.
When the original answer differs from the most frequent negated answer,
SFT raises answer change from 53.2\% to 94.8\%. However, when the two
match, answer change falls from 34.2\% to 18.8\%.
In contrast, ALiT improves both groups, to 67.1\% and 40.2\%,
respectively. Thus, ALiT improves negation without strongly amplifying
the negation bias identified in \S~\ref{sec:evaluation}.


\section{Conclusion}
We investigate why LLMs can answer a question correctly yet fail to
reject that answer under negation. We identify attention heads and
MLP neurons that suppress original-answer retrieval while preserving
answer category information and promoting favored candidates.
However, these candidate preferences do not reliably adjust to which
answer must be excluded, leaving negation vulnerable to insufficient
correction and negation bias.
Guided by this diagnosis, we propose ALiT, which requires stronger
corrections for more confident original predictions. ALiT improves
negation while largely preserving general performance and avoiding
the sharp loss of answer diversity observed with SFT.
Our results demonstrate how mechanistic analysis can explain linguistic
failures and guide training that addresses their underlying limitations.



\subsection*{AI use statement}

In this work, we used generative AI tools to provide feedback on research methodology and experimental design, assist in the writing of proofs and translation, support qualitative and thematic data analysis, and assist with the interpretation of experimental results. We also used an LLM-as-a-Judge for benchmark validation and model evaluation, as described in the corresponding sections. We did not use generative AI tools to generate synthetic datasets, develop theoretical models or conceptual frameworks, formulate mathematical claims or provide critical ingredients for proving them, propose or refine hypotheses, implement methods, or clean or reformat datasets. Additionally, we used generative AI tools to identify potentially relevant literature, improve the clarity and grammar of the manuscript, assist with LaTeX code, and support the preparation of figures. All experimental designs and research decisions were independently reviewed and validated by the authors, and relevant literature suggested by generative AI tools was manually verified against the original sources. AI-assisted work was reviewed and revised by the authors for correctness and consistency with the underlying experiments and results. We take responsibility for the final content of this work, including text, claims, and artifacts produced with the aid of generative AI.




\subsection*{Reproducibility statement}
We provide a detailed description of our benchmark construction and validation procedures in Section~\ref{sec:evaluation} and Appendix~\ref{app:benchmark}, including the source benchmarks, negation transformation rules, automatic checks, and LLM- and human-based validation procedures. We further describe the model evaluation procedure and metrics in Appendix~\ref{app:evaluation_details}.
The procedures and settings for our mechanistic analyses, including information-flow tracing, negation-signal interventions, component identification, and subspace experiments, are described in Appendix~\ref{app:experiment_details}, with additional results in Appendix~\ref{app:additional-results}. Section~\ref{sec:negation-training} specifies our proposed training objective, and Appendix~\ref{app:training} provides the complete definitions of ALiT and the baselines, target construction procedures, optimization settings, and training hyperparameters.




\bibliography{iclr2027_conference}

@inproceedings{truong2023naysayers,
  title     = {Language Models Are Not Naysayers: An Analysis of Language Models on Negation Benchmarks},
  author    = {Truong, Thinh Hung and Baldwin, Timothy and Verspoor, Karin and Cohn, Trevor},
  booktitle = {Proceedings of the 12th Joint Conference on Lexical and Computational Semantics (*SEM 2023)},
  pages     = {101--114},
  year      = {2023},
  address   = {Toronto, Canada},
  publisher = {Association for Computational Linguistics},
  doi       = {10.18653/v1/2023.starsem-1.10},
  url       = {https://aclanthology.org/2023.starsem-1.10/}
}

@inproceedings{garciaferrero2023notdataset,
  title     = {This is not a Dataset: A Large Negation Benchmark to Challenge Large Language Models},
  author    = {Garc{\'i}a-Ferrero, Iker and Altuna, Bego{\~n}a and Alvez, Javier and Gonzalez-Dios, Itziar and Rigau, German},
  booktitle = {Proceedings of the 2023 Conference on Empirical Methods in Natural Language Processing},
  pages     = {8596--8615},
  year      = {2023},
  address   = {Singapore},
  publisher = {Association for Computational Linguistics},
  doi       = {10.18653/v1/2023.emnlp-main.531},
  url       = {https://aclanthology.org/2023.emnlp-main.531/}
}

@inproceedings{zhang2023beyond,
  title     = {Beyond Positive Scaling: How Negation Impacts Scaling Trends of Language Models},
  author    = {Zhang, Yuhui and Yasunaga, Michihiro and Zhou, Zhengping and HaoChen, Jeff Z. and Zou, James and Liang, Percy and Yeung, Serena},
  booktitle = {Findings of the Association for Computational Linguistics: ACL 2023},
  pages     = {7479--7498},
  year      = {2023},
  address   = {Toronto, Canada},
  publisher = {Association for Computational Linguistics},
  doi       = {10.18653/v1/2023.findings-acl.472},
  url       = {https://aclanthology.org/2023.findings-acl.472/}
}

@inproceedings{barreto2025disimprovement,
  title     = {This is not a Disimprovement: Improving Negation Reasoning in Large Language Models via Prompt Engineering},
  author    = {Barreto, Joshua Jose Dias and Jana, Abhik},
  booktitle = {Findings of the Association for Computational Linguistics: EMNLP 2025},
  pages     = {14149--14156},
  year      = {2025},
  publisher = {Association for Computational Linguistics},
  url       = {https://aclanthology.org/2025.findings-emnlp.761/}
}

@inproceedings{
meng2022locating,
title={Locating and Editing Factual Associations in {GPT}},
author={Kevin Meng and David Bau and Alex J Andonian and Yonatan Belinkov},
booktitle={Advances in Neural Information Processing Systems},
editor={Alice H. Oh and Alekh Agarwal and Danielle Belgrave and Kyunghyun Cho},
year={2022},
url={https://openreview.net/forum?id=-h6WAS6eE4}
}

@article{warstadt-etal-2020-blimp-benchmark,
    title = "{BL}i{MP}: The Benchmark of Linguistic Minimal Pairs for {E}nglish",
    author = "Warstadt, Alex  and
      Parrish, Alicia  and
      Liu, Haokun  and
      Mohananey, Anhad  and
      Peng, Wei  and
      Wang, Sheng-Fu  and
      Bowman, Samuel R.",
    editor = "Johnson, Mark  and
      Roark, Brian  and
      Nenkova, Ani",
    journal = "Transactions of the Association for Computational Linguistics",
    volume = "8",
    year = "2020",
    address = "Cambridge, MA",
    publisher = "MIT Press",
    url = "https://aclanthology.org/2020.tacl-1.25/",
    doi = "10.1162/tacl_a_00321",
    pages = "377--392"
}

@inproceedings{marvin-linzen-2018-targeted,
    title = "Targeted Syntactic Evaluation of Language Models",
    author = "Marvin, Rebecca  and
      Linzen, Tal",
    editor = "Riloff, Ellen  and
      Chiang, David  and
      Hockenmaier, Julia  and
      Tsujii, Jun{'}ichi",
    booktitle = "Proceedings of the 2018 Conference on Empirical Methods in Natural Language Processing",
    month = oct # "-" # nov,
    year = "2018",
    address = "Brussels, Belgium",
    publisher = "Association for Computational Linguistics",
    url = "https://aclanthology.org/D18-1151/",
    doi = "10.18653/v1/D18-1151",
    pages = "1192--1202"
}

@inproceedings{hartmann-etal-2021-multilingual,
    title = "A Multilingual Benchmark for Probing Negation-Awareness with Minimal Pairs",
    author = "Hartmann, Mareike  and
      de Lhoneux, Miryam  and
      Hershcovich, Daniel  and
      Kementchedjhieva, Yova  and
      Nielsen, Lukas  and
      Qiu, Chen  and
      S{\o}gaard, Anders",
    editor = "Bisazza, Arianna  and
      Abend, Omri",
    booktitle = "Proceedings of the 25th Conference on Computational Natural Language Learning",
    month = nov,
    year = "2021",
    address = "Online",
    publisher = "Association for Computational Linguistics",
    url = "https://aclanthology.org/2021.conll-1.19/",
    doi = "10.18653/v1/2021.conll-1.19",
    pages = "244--257"
}

@inproceedings{negation_gpt2,
   title={Interpreting Negation in GPT-2: Layer-and Head-Level Causal Analysis},
   url={http://dx.doi.org/10.1109/CCWC67433.2026.11393646},
   DOI={10.1109/ccwc67433.2026.11393646},
   booktitle={2026 IEEE 16th Annual Computing and Communication Workshop and Conference (CCWC)},
   publisher={IEEE},
   author={Al Mofael, Abdullah and Kuhn, Lisa M. and Alkadi, Ghassan and Yang, Kuo-Pao},
   year={2026},
   month=Jan, pages={0042–0050} }

@inproceedings{hierarchical_concepts,
 author = {Park, Kiho and Choe, Yo Joong and Jiang, Yibo and Veitch, Victor},
 booktitle = {International Conference on Learning Representations},
 editor = {Y. Yue and A. Garg and N. Peng and F. Sha and R. Yu},
 pages = {76441--76463},
 title = {The Geometry of Categorical and Hierarchical Concepts in Large Language Models},
 url = {https://proceedings.iclr.cc/paper_files/paper/2025/file/be7430d22a4dae8516894e32f2fcc6db-Paper-Conference.pdf},
 volume = {2025},
 year = {2025}
}

@inproceedings{actpatch,
 author = {Zhang, Fred and Nanda, Neel},
 booktitle = {International Conference on Learning Representations},
 editor = {B. Kim and Y. Yue and S. Chaudhuri and K. Fragkiadaki and M. Khan and Y. Sun},
 pages = {1651--1678},
 title = {Towards Best Practices of Activation Patching in Language Models: Metrics and Methods},
 url = {https://proceedings.iclr.cc/paper_files/paper/2024/file/06a52a54c8ee03cd86771136bc91eb1f-Paper-Conference.pdf},
 volume = {2024},
 year = {2024}
}

@inproceedings{PTR,
 author = {Hong, Yining and Yi, Li and Tenenbaum, Josh and Torralba, Antonio and Gan, Chuang},
 booktitle = {Advances in Neural Information Processing Systems},
 editor = {M. Ranzato and A. Beygelzimer and Y. Dauphin and P.S. Liang and J. Wortman Vaughan},
 pages = {17427--17440},
 publisher = {Curran Associates, Inc.},
 title = {PTR: A Benchmark for Part-based Conceptual, Relational, and Physical Reasoning},
 url = {https://proceedings.neurips.cc/paper_files/paper/2021/file/918f5cd5a5c0d48671d4d4fc54bab2e9-Paper.pdf},
 volume = {34},
 year = {2021}
}

@INPROCEEDINGS{GQA,
  author={Hudson, Drew A. and Manning, Christopher D.},
  booktitle={2019 IEEE/CVF Conference on Computer Vision and Pattern Recognition (CVPR)}, 
  title={GQA: A New Dataset for Real-World Visual Reasoning and Compositional Question Answering}, 
  year={2019},
  volume={},
  number={},
  pages={6693-6702},
  doi={10.1109/CVPR.2019.00686}}

@inproceedings{synthworld,
 author = {Gu, Ken and Bhat, Advait and Merrill, Mike and West, Robert and Liu, Xin and McDuff, Daniel and Althoff, Tim},
 booktitle = {International Conference on Learning Representations},
 editor = {C. Vondrick and B. Hariharan and C. Raffel and L. Pinto and D. Yang and A. Faust},
 pages = {109525--109573},
 title = {SynthWorlds: Controlled Parallel Worlds for Disentangling Reasoning and Knowledge in Language Models},
 url = {https://proceedings.iclr.cc/paper_files/paper/2026/file/b213d870740582dd6af77bbdaed900c9-Paper-Conference.pdf},
 volume = {2026},
 year = {2026}
}

@inproceedings{
phantomwiki,
title={PhantomWiki: On-Demand Datasets for Reasoning and Retrieval Evaluation},
author={Albert Gong and Kamil{\.{e}} Stankevi{\v{c}}i{\={u}}t{\.{e}} and Chao Wan and Anmol Kabra and Raphael Thesmar and Johann Lee and Julius Klenke and Carla P Gomes and Kilian Q Weinberger},
booktitle={Forty-second International Conference on Machine Learning},
year={2025},
url={https://openreview.net/forum?id=DIZItj8ueN}
}

@article{rippleedit,
    title = "Evaluating the Ripple Effects of Knowledge Editing in Language Models",
    author = "Cohen, Roi  and
      Biran, Eden  and
      Yoran, Ori  and
      Globerson, Amir  and
      Geva, Mor",
    journal = "Transactions of the Association for Computational Linguistics",
    volume = "12",
    year = "2024",
    address = "Cambridge, MA",
    publisher = "MIT Press",
    url = "https://aclanthology.org/2024.tacl-1.16/",
    doi = "10.1162/tacl_a_00644",
    pages = "283--298"
}

@inproceedings{popqa,
    title = "When Not to Trust Language Models: Investigating Effectiveness of Parametric and Non-Parametric Memories",
    author = "Mallen, Alex  and
      Asai, Akari  and
      Zhong, Victor  and
      Das, Rajarshi  and
      Khashabi, Daniel  and
      Hajishirzi, Hannaneh",
    editor = "Rogers, Anna  and
      Boyd-Graber, Jordan  and
      Okazaki, Naoaki",
    booktitle = "Proceedings of the 61st Annual Meeting of the Association for Computational Linguistics (Volume 1: Long Papers)",
    month = jul,
    year = "2023",
    address = "Toronto, Canada",
    publisher = "Association for Computational Linguistics",
    url = "https://aclanthology.org/2023.acl-long.546/",
    doi = "10.18653/v1/2023.acl-long.546",
    pages = "9802--9822"
}

@misc{nostalgebraist2020logitlens,
  author       = {Nostalgebraist},
  title        = {Interpreting GPT: The Logit Lens},
  howpublished = {LessWrong},
  year         = {2020},
  url          = {https://www.lesswrong.com/posts/AcKRB8wDpdaN6v6ru/interpreting-gpt-the-logit-lens}
}

@inproceedings{jang2023negated,
  title     = {Can Large Language Models Truly Understand Prompts? A Case Study with Negated Prompts},
  author    = {Jang, Joel and Ye, Seonghyeon and Seo, Minjoon},
  booktitle = {Proceedings of the 1st Transfer Learning for Natural Language Processing Workshop},
  pages     = {52--62},
  year      = {2023},
  volume    = {203},
  series    = {Proceedings of Machine Learning Research},
  publisher = {PMLR},
  url       = {https://proceedings.mlr.press/v203/jang23a.html}
}

@article{rafailov2023direct,
  title={Direct preference optimization: Your language model is secretly a reward model},
  author={Rafailov, Rafael and Sharma, Archit and Mitchell, Eric and Manning, Christopher D and Ermon, Stefano and Finn, Chelsea},
  journal={Advances in neural information processing systems},
  volume={36},
  pages={53728--53741},
  year={2023}
}

@InProceedings{DAS,
  title = 	 {Finding Alignments Between Interpretable Causal Variables and Distributed Neural Representations},
  author =       {Geiger, Atticus and Wu, Zhengxuan and Potts, Christopher and Icard, Thomas and Goodman, Noah},
  booktitle = 	 {Proceedings of the Third Conference on Causal Learning and Reasoning},
  pages = 	 {160--187},
  year = 	 {2024},
  editor = 	 {Locatello, Francesco and Didelez, Vanessa},
  volume = 	 {236},
  series = 	 {Proceedings of Machine Learning Research},
  month = 	 {01--03 Apr},
  publisher =    {PMLR},
  url = 	 {https://proceedings.mlr.press/v236/geiger24a.html}
}

@inproceedings{
lrh,
title={The Linear Representation Hypothesis and the Geometry of Large Language Models},
author={Kiho Park and Yo Joong Choe and Victor Veitch},
booktitle={Forty-first International Conference on Machine Learning},
year={2024},
url={https://openreview.net/forum?id=UGpGkLzwpP}
}

@inproceedings{
zhou2026how,
title={How Language Models Process Negation},
author={Zhejian Zhou and Tianyi Zhou and Robin Jia and Jonathan May},
booktitle={Forty-third International Conference on Machine Learning},
year={2026},
url={https://openreview.net/forum?id=8DLW34pkNi}
}

@inproceedings{ravichander-etal-2022-condaqa,
    title = "{CONDAQA}: A Contrastive Reading Comprehension Dataset for Reasoning about Negation",
    author = "Ravichander, Abhilasha  and
      Gardner, Matt  and
      Marasovic, Ana",
    editor = "Goldberg, Yoav  and
      Kozareva, Zornitsa  and
      Zhang, Yue",
    booktitle = "Proceedings of the 2022 Conference on Empirical Methods in Natural Language Processing",
    month = dec,
    year = "2022",
    address = "Abu Dhabi, United Arab Emirates",
    publisher = "Association for Computational Linguistics",
    url = "https://aclanthology.org/2022.emnlp-main.598/",
    doi = "10.18653/v1/2022.emnlp-main.598",
    pages = "8729--8755"
}

@inproceedings{she-etal-2023-scone,
    title = "{S}co{N}e: Benchmarking Negation Reasoning in Language Models With Fine-Tuning and In-Context Learning",
    author = "She, Jingyuan S.  and
      Potts, Christopher  and
      Bowman, Samuel R.  and
      Geiger, Atticus",
    editor = "Rogers, Anna  and
      Boyd-Graber, Jordan  and
      Okazaki, Naoaki",
    booktitle = "Proceedings of the 61st Annual Meeting of the Association for Computational Linguistics (Volume 2: Short Papers)",
    month = jul,
    year = "2023",
    address = "Toronto, Canada",
    publisher = "Association for Computational Linguistics",
    url = "https://aclanthology.org/2023.acl-short.154/",
    doi = "10.18653/v1/2023.acl-short.154",
    pages = "1803--1821"
}

@article{ettinger-2020-bert,
    title = "What {BERT} Is Not: Lessons from a New Suite of Psycholinguistic Diagnostics for Language Models",
    author = "Ettinger, Allyson",
    editor = "Johnson, Mark  and
      Roark, Brian  and
      Nenkova, Ani",
    journal = "Transactions of the Association for Computational Linguistics",
    volume = "8",
    year = "2020",
    address = "Cambridge, MA",
    publisher = "MIT Press",
    url = "https://aclanthology.org/2020.tacl-1.3/",
    doi = "10.1162/tacl_a_00298",
    pages = "34--48"
}

@inproceedings{kassner-schutze-2020-negated,
    title = "Negated and Misprimed Probes for Pretrained Language Models: Birds Can Talk, But Cannot Fly",
    author = {Kassner, Nora  and
      Sch{\"u}tze, Hinrich},
    editor = "Jurafsky, Dan  and
      Chai, Joyce  and
      Schluter, Natalie  and
      Tetreault, Joel",
    booktitle = "Proceedings of the 58th Annual Meeting of the Association for Computational Linguistics",
    month = jul,
    year = "2020",
    address = "Online",
    publisher = "Association for Computational Linguistics",
    url = "https://aclanthology.org/2020.acl-main.698/",
    doi = "10.18653/v1/2020.acl-main.698",
    pages = "7811--7818"
}

@article{zuanazzi2024negation,
  title={Negation mitigates rather than inverts the neural representations of adjectives},
  author={Zuanazzi, Arianna and Ripoll{\'e}s, Pablo and Lin, Wy Ming and Gwilliams, Laura and King, Jean-R{\'e}mi and Poeppel, David},
  journal={PLoS biology},
  volume={22},
  number={5},
  pages={e3002622},
  year={2024},
  publisher={Public Library of Science San Francisco, CA USA}
}

@inproceedings{geva-etal-2023-dissecting,
  title     = {Dissecting Recall of Factual Associations in Auto-Regressive Language Models},
  author    = {Geva, Mor and Bastings, Jasmijn and Filippova, Katja and Globerson, Amir},
  booktitle = {Proceedings of the 2023 Conference on Empirical Methods in Natural Language Processing},
  pages     = {12216--12235},
  year      = {2023},
  month     = dec,
  address   = {Singapore},
  publisher = {Association for Computational Linguistics},
  doi       = {10.18653/v1/2023.emnlp-main.751},
  url       = {https://aclanthology.org/2023.emnlp-main.751/}
}

@article{fischler1983brain,
  title   = {Brain Potentials Related to Stages of Sentence Verification},
  author  = {Fischler, Ira and Bloom, Paul A. and Childers, Donald G. and Roucos, Salim E. and Perry, Jr., Nathan W.},
  journal = {Psychophysiology},
  volume  = {20},
  number  = {4},
  pages   = {400--409},
  year    = {1983},
  month   = jul,
  doi     = {10.1111/j.1469-8986.1983.tb00920.x},
  url     = {https://doi.org/10.1111/j.1469-8986.1983.tb00920.x}
}

@article{mayo2004successful,
  title   = {{``I am not guilty'' vs. ``I am innocent'': Successful negation may depend on the schema used for its encoding}},
  author  = {Mayo, Ruth and Schul, Yaacov and Burnstein, Eugene},
  journal = {Journal of Experimental Social Psychology},
  volume  = {40},
  number  = {4},
  pages   = {433--449},
  year    = {2004},
  month   = jul,
  doi     = {10.1016/j.jesp.2003.07.008},
  url     = {https://doi.org/10.1016/j.jesp.2003.07.008}
}

@article{nieuwland2008truth,
  title   = {When the Truth Is Not Too Hard to Handle: An Event-Related Potential Study on the Pragmatics of Negation},
  author  = {Nieuwland, Mante S. and Kuperberg, Gina R.},
  journal = {Psychological Science},
  volume  = {19},
  number  = {12},
  pages   = {1213--1218},
  year    = {2008},
  month   = dec,
  doi     = {10.1111/j.1467-9280.2008.02226.x},
  url     = {https://doi.org/10.1111/j.1467-9280.2008.02226.x}
}

@article{clark1972process,
  title   = {On the Process of Comparing Sentences Against Pictures},
  author  = {Clark, Herbert H. and Chase, William G.},
  journal = {Cognitive Psychology},
  volume  = {3},
  number  = {3},
  pages   = {472--517},
  year    = {1972},
  doi     = {10.1016/0010-0285(72)90019-9},
  url     = {https://doi.org/10.1016/0010-0285(72)90019-9}
}

@inproceedings{hosseini-etal-2021-understanding,
    title = "Understanding by Understanding Not: Modeling Negation in Language Models",
    author = "Hosseini, Arian and
      Reddy, Siva and
      Bahdanau, Dzmitry and
      Hjelm, R Devon and
      Sordoni, Alessandro and
      Courville, Aaron",
    booktitle = "Proceedings of the 2021 Conference of the North American Chapter of the Association for Computational Linguistics: Human Language Technologies",
    month = jun,
    year = "2021",
    address = "Online",
    publisher = "Association for Computational Linguistics",
    url = "https://aclanthology.org/2021.naacl-main.102/",
    doi = "10.18653/v1/2021.naacl-main.102",
    pages = "1301--1312"
}

@inproceedings{helwe-etal-2022-tina,
    title = "{TINA}: Textual Inference with Negation Augmentation",
    author = "Helwe, Chadi and
      Coumes, Simon and
      Clavel, Chlo{\'e} and
      Suchanek, Fabian",
    booktitle = "Findings of the Association for Computational Linguistics: EMNLP 2022",
    month = dec,
    year = "2022",
    address = "Abu Dhabi, United Arab Emirates",
    publisher = "Association for Computational Linguistics",
    url = "https://aclanthology.org/2022.findings-emnlp.301/",
    doi = "10.18653/v1/2022.findings-emnlp.301",
    pages = "4086--4099"
}

@inproceedings{quantmeyer-etal-2024-clip,
    title = "How and where does {CLIP} process negation?",
    author = "Quantmeyer, Vincent and
      Mosteiro, Pablo and
      Gatt, Albert",
    booktitle = "Proceedings of the 3rd Workshop on Advances in Language and Vision Research (ALVR)",
    month = aug,
    year = "2024",
    address = "Bangkok, Thailand",
    publisher = "Association for Computational Linguistics",
    url = "https://aclanthology.org/2024.alvr-1.5/",
    doi = "10.18653/v1/2024.alvr-1.5",
    pages = "59--72"
}

@inproceedings{Alhamoud_2025_CVPR,
    author = {Alhamoud, Kumail and Alshammari, Shaden and Tian, Yonglong and
              Li, Guohao and Torr, Philip H.S. and Kim, Yoon and Ghassemi, Marzyeh},
    title = {Vision-Language Models Do Not Understand Negation},
    booktitle = {Proceedings of the IEEE/CVF Conference on Computer Vision and Pattern Recognition (CVPR)},
    month = {June},
    year = {2025},
    pages = {29612--29622}
}

@inproceedings{syed-etal-2024-attribution,
    title = "Attribution Patching Outperforms Automated Circuit Discovery",
    author = "Syed, Aaquib  and
      Rager, Can  and
      Conmy, Arthur",
    booktitle = "Proceedings of the 7th BlackboxNLP Workshop: Analyzing and Interpreting Neural Networks for NLP",
    month = nov,
    year = "2024",
    address = "Miami, Florida, US",
    publisher = "Association for Computational Linguistics",
    url = "https://aclanthology.org/2024.blackboxnlp-1.25/",
    doi = "10.18653/v1/2024.blackboxnlp-1.25",
    pages = "407--416",
}
\bibliographystyle{iclr2027_conference}

\clearpage
\appendix

\addtocontents{toc}{\protect\setcounter{tocdepth}{2}}

\begingroup
\renewcommand{\contentsname}{Appendix Contents}
\tableofcontents
\endgroup

\clearpage


\section{Related Work}

\subsection{Negation Processing in Large Language Models and Humans}


Language models have long been known to struggle with negation, including masked and encoder-based models \citep{ettinger-2020-bert,kassner-schutze-2020-negated}. These difficulties persist in generative LLMs across language understanding, reasoning, and factual prediction \citep{truong2023naysayers,garciaferrero2023notdataset,zhang2023beyond}. 
Prior work has explored remedies, including training objectives, data augmentation, fine-tuning, and prompting \citep{hosseini-etal-2021-understanding,helwe-etal-2022-tina,she-etal-2023-scone,jang2023negated,barreto2025disimprovement}.
Negation failures also extend to vision-language models, where CLIP-based models struggle to distinguish negated descriptions \citep{Alhamoud_2025_CVPR, quantmeyer-etal-2024-clip}.
More recently, mechanistic studies have begun to investigate how language models internally process negation \citep{negation_gpt2,zhou2026how}.

Most closely related to our work, \cite{zhou2026how} mechanistically study
negation in a controlled family of templated prompts, identifying suppression
and construction mechanisms for representing negated concepts. They further
show that these mechanisms can be overridden by late-layer shortcut attention,
attributing failures primarily to competition from a separate mechanism.
Our work examines both processes at a finer granularity across diverse domains
and negation forms. For suppression, whereas \cite{zhou2026how} characterize how
negation-related modules reduce support for the negated concept, we trace the
underlying retrieval change: attention heads that support the original answer
in the non-negated prompt reduce their attention to answer tokens under
negation. For construction, we go beyond identifying a constructed negated
representation and trace how it produces a concrete alternative, separating
the preservation of answer category information from shifts in candidate
preference. This analysis reveals an intrinsic source of negation failure:
because the model relies on suppressing the original answer and promoting
alternatives, the correction itself can be insufficient or favor the original
answer. We use this mechanistic diagnosis to derive a confidence-adaptive
training objective.


Classical accounts of human negation propose that people first represent
the non-negated content and then verify or revise it to understand the
negated statement
\citep{clark1972process,fischler1983brain,mayo2004successful}.
Later work suggests that these processes need not be strictly sequential:
negation interacts with context and available alternatives as the meaning
of the statement is being formed
\citep{nieuwland2008truth,zuanazzi2024negation}.
These accounts motivate us to examine how LLMs use the information
supporting the original answer under negation, without assuming that they
follow the same mechanism. Specifically, we ask whether models replace
the original representation or retain and revise it, and whether this
revision occurs only after answer formation or during the computation
that produces the answer.

\subsection{Mechanistic Interpretability}

Causal studies of factual recall identify components that retrieve and
transmit information supporting an answer
\citep{meng2022locating,geva-etal-2023-dissecting}.
Activation and attribution patching provide complementary tools for tracing
information flow and identifying components that causally contribute to model
behavior \citep{actpatch,syed-etal-2024-attribution}.
We build on these approaches to trace how negation reaches the prediction
position and to identify attention heads and MLP neurons that suppress the
original answer, promote answer category information, and favor particular
alternatives.
The linear representation hypothesis and work on categorical concepts
motivate distinguishing answer category information from candidate preference
\citep{lrh,hierarchical_concepts}.
However, geometric separation alone does not establish different causal roles.
We therefore use DAS \citep{DAS} to learn intervention-defined subspaces and
test their causal roles on held-out predictions.

\section{Details on the Construction of the Negation Benchmark}
\label{app:benchmark}



\subsection{Benchmark Construction}
\label{app:benchmark_construction}

We construct paired prompts with and without negation from six source benchmarks covering factual association, logical reasoning, and visual question answering (\S~\ref{sec:evaluation}). This rule-based construction follows a minimal-pair design \citep{marvin-linzen-2018-targeted,warstadt-etal-2020-blimp-benchmark,hartmann-etal-2021-multilingual}. 
Each source example provides a question or completion prompt, reference answer(s), and, where applicable, supporting text or an image. Our construction procedure combines source-provided task structure, syntactic analysis, and deterministic transformation rules. Table~\ref{tab:benchmark_examples} shows all four negation forms for one example from each benchmark.

\paragraph{Source selection.}
We use selected portions of the six source benchmarks rather than
all available examples. We retain questions that can be converted
into controlled negation pairs while preserving their supporting
context or image. The source collections and benchmark-specific
selection rules are as follows.

\begin{itemize}[leftmargin=1.5em]
    \item \textbf{PopQA.}
    We start from 14,267 questions and exclude records with missing
    questions, subjects, relations, or reference answers.
    We further exclude questions whose sentence structure is
    unsupported by our transformation rules or whose negation scope
    cannot be determined reliably.

    \item \textbf{RippleEdits.}
    We use the \texttt{popular}, \texttt{random}, and \texttt{recent}
    collections to construct completion prompts from recoverable
    source facts and answer targets. We exclude records for which
    these cannot be reconstructed, rather than using every
    evaluation question associated with a knowledge edit.

    \item \textbf{PhantomWiki.}
    We generate 20 worlds with depth 20, size 50, and seeds
    1--20, with ten questions per template.
    We exclude aggregate count questions and retain only questions
    with exactly one distinct answer after normalization and
    deduplication. This avoids treating another valid answer to the
    original question as an alternative under negation.

    \item \textbf{SynthWorlds.}
    We use the 1,200 examples in the SM subset.
    We exclude questions with ambiguous negation scope, unsupported
    sentence structures, or conjunctions that would require
    rewriting the task rather than negating the queried relation.

    \item \textbf{GQA.}
    We start from the 12,578 questions in the balanced test-dev
    collection and retain questions with the structural type
    \texttt{query}. We then apply our transformation and scope
    checks, excluding questions that do not support the required
    controlled contrast.

    \item \textbf{PTR.}
    We start from the 91,720 validation questions.
    Within each scene, we identify open-ended questions supported
    by our transformation rules and uniformly sample one eligible
    question using the fixed seed 20260909.
    Scenes without an eligible question are excluded.
\end{itemize}

The upstream collection names, such as GQA test-dev and PTR
validation, do not refer to the splits of our benchmark.
We construct our own train, validation, and test partition, keeping
the negation variants of each source case together.

\paragraph{Pair-level filtering.}
After source selection and automatic construction checks, we
evaluate each negation variant separately for grammaticality,
unambiguity, and negation scope fidelity
(\S~\ref{app:benchmark_validation}).
A source question can therefore retain fewer than four variants.
We also remove the instruction-based \texttt{incorrect\_option}
variant, retaining only the four linguistic negation forms.
Benchmark inclusion does not depend on whether a model answers
the question correctly.

This procedure retains 105,048 pairs from 126,756 constructed
linguistic pairs. The reviewed release preserves the existing
split assignments and contains 41,957 training pairs, 10,348
validation pairs, and 52,743 test pairs.
These counts describe the benchmark splits, not the smaller
cohorts selected for individual analyses or post-training.
\paragraph{Identifying the relation to negate.}
We first identify the part of the question that asks for the answer, rather than inserting negation before an arbitrary verb. Where available, we use the source benchmark's structured representation: the outer Prolog predicate in PhantomWiki, the final decomposed question in SynthWorlds, the terminal semantic query in GQA, and the final program operation in PTR. These structures distinguish the relation being queried from intermediate relations used to identify an entity. For RippleEdits, the completion boundary identifies the relation whose answer is missing. For PopQA, we use the question structure and the provided subject span. Supporting contexts and images remain unchanged.

\paragraph{Syntactic analysis.}
We analyze the question using spaCy~3.8.14 with the English model \texttt{en\_core\_web\_sm}~3.8.0. Its pipeline provides tokenization, part-of-speech tags, lemmas, and dependency parses. We use these annotations to locate the relevant predicate and its auxiliary, determine the required verb form, and distinguish the main clause from relative clauses or other modifiers. Thus, a verb inside a phrase identifying the subject is not automatically treated as the predicate to negate.

\paragraph{Transformation rules.}
We apply a fixed set of grammatical rules to the selected predicate:
\begin{itemize}[leftmargin=1.5em]
    \item \textit{Not.} If an auxiliary or copula is already present, we insert \textit{not} at the corresponding predicate boundary. Otherwise, we introduce \textit{do}, \textit{does}, or \textit{did}, preserving tense and agreement, and use the base form of the lexical verb. Verb forms are determined from the syntactic annotations and predefined morphological rules.
    \item \textit{Contracted negation.} We replace the relevant auxiliary with its registered contraction, such as \textit{isn't}, \textit{doesn't}, or \textit{didn't}, using the corresponding question or completion construction.
    \item \textit{Never} and \textit{by no means.} We insert the expression at the predicate boundary identified for canonical \textit{not}. When \textit{not} requires newly introduced do-support, these adverbial forms can instead modify the original inflected verb directly; for example, \textit{did not influence} contrasts with \textit{never influenced}.
\end{itemize}

Some source questions require grammatical normalization before these rules can be applied. For example, attribute questions can be expressed as ``What is the color of \ldots?'', and a subject question such as ``Who influenced \ldots?'' becomes ``Who did influence \ldots?'' for the \textit{not} and contracted pairs. We apply the same normalization to both prompts within each pair. The non-negated prompt can therefore differ across forms, as illustrated by the SynthWorlds example in Table~\ref{tab:benchmark_examples}.

\paragraph{Automatic checks and subsequent validation.}
We check that each pair differs only by its registered insertion or contraction and preserves the remaining input, including the registered subject span. Dependency-based checks verify that negation targets the intended predicate rather than a retained modifier, and that relative-clause text remains unchanged. The adverbial forms reuse the canonical scope decision and undergo exact insertion checks. For predefined structural templates, source-provided semantic structure and exact edit checks can resolve parser attachment errors; otherwise, ambiguous or unsupported constructions are excluded.

These checks establish how the edit was made, but do not guarantee semantic validity. In particular, \textit{never} can introduce a temporal interpretation. We therefore validate each generated form separately using the procedure in \S~\ref{app:benchmark_validation}. A retained question need not have all four variants. The examples below are selected to illustrate all four forms, not to represent their relative frequencies.

\begingroup
\small
\setlength{\tabcolsep}{4pt}
\renewcommand{\arraystretch}{1.12}
\begin{longtable}{@{}
    >{\raggedright\arraybackslash}p{0.19\linewidth}
    >{\raggedright\arraybackslash}p{0.76\linewidth}
@{}}
\caption{Examples of all four negation forms from each source benchmark, drawn from the training split. Bold text marks the negation expression. Original answers are source references, not targets for the negated prompts. Underlines indicate completion slots for display. Supporting documents are omitted; the corresponding GQA and PTR images appear in Figure~\ref{fig:benchmark_example_images}.}
\label{tab:benchmark_examples}\\
\toprule
Form & Prompt \\
\midrule
\endfirsthead

\caption[]{Examples of benchmark construction, continued.}\\
\toprule
Form & Prompt \\
\midrule
\endhead

\bottomrule
\endfoot

\multicolumn{2}{@{}p{0.95\linewidth}@{}}{
\textbf{PopQA}\quad Original answer: New Delhi
}\\*
Original
& What is the capital of India? \\*
\textit{not}
& What is \textbf{not} the capital of India? \\*
\textit{n't}
& What \textbf{isn't} the capital of India? \\*
\textit{never}
& What is \textbf{never} the capital of India? \\*
\textit{by no means}
& What is \textbf{by no means} the capital of India? \\
\midrule

\multicolumn{2}{@{}p{0.95\linewidth}@{}}{
\textbf{RippleEdits}\quad Original answer: association football player
}\\*
Original
& The occupation of Enrico Chiesa is \underline{\hspace{0.4cm}} \\*
\textit{not}
& The occupation of Enrico Chiesa is \textbf{not} \underline{\hspace{0.4cm}} \\*
\textit{n't}
& The occupation of Enrico Chiesa \textbf{isn't} \underline{\hspace{0.4cm}} \\*
\textit{never}
& The occupation of Enrico Chiesa is \textbf{never} \underline{\hspace{0.4cm}} \\*
\textit{by no means}
& The occupation of Enrico Chiesa is \textbf{by no means} \underline{\hspace{0.4cm}} \\
\midrule

\multicolumn{2}{@{}p{0.95\linewidth}@{}}{
\textbf{PhantomWiki}\quad Original answer: Stewart Herrera
}\\*
Original
& Who is the father-in-law of Vita Herrera? \\*
\textit{not}
& Who is \textbf{not} the father-in-law of Vita Herrera? \\*
\textit{n't}
& Who \textbf{isn't} the father-in-law of Vita Herrera? \\*
\textit{never}
& Who is \textbf{never} the father-in-law of Vita Herrera? \\*
\textit{by no means}
& Who is \textbf{by no means} the father-in-law of Vita Herrera? \\
\midrule

\multicolumn{2}{@{}p{0.95\linewidth}@{}}{
\textbf{SynthWorlds}\quad Original answer: Thalric Kytarathian Stormrider
}\\*
Original
& Who influenced the field of work of Vail Thorne? \\*
Original for \textit{not}/\textit{n't}
& Who did influence the field of work of Vail Thorne? \\*
\textit{not}
& Who did \textbf{not} influence the field of work of Vail Thorne? \\*
\textit{n't}
& Who \textbf{didn't} influence the field of work of Vail Thorne? \\*
\textit{never}
& Who \textbf{never} influenced the field of work of Vail Thorne? \\*
\textit{by no means}
& Who \textbf{by no means} influenced the field of work of Vail Thorne? \\
\midrule

\multicolumn{2}{@{}p{0.95\linewidth}@{}}{
\textbf{GQA}\quad Original answer: blue
}\\*
Original
& What is the color of the clean shirt? \\*
\textit{not}
& What is \textbf{not} the color of the clean shirt? \\*
\textit{n't}
& What \textbf{isn't} the color of the clean shirt? \\*
\textit{never}
& What is \textbf{never} the color of the clean shirt? \\*
\textit{by no means}
& What is \textbf{by no means} the color of the clean shirt? \\
\midrule

\multicolumn{2}{@{}p{0.95\linewidth}@{}}{
\textbf{PTR}\quad Original answer: blue
}\\*
Original
& what is the color of the sleep area of the bed ? \\*
\textit{not}
& what is \textbf{not} the color of the sleep area of the bed ? \\*
\textit{n't}
& what \textbf{isn't} the color of the sleep area of the bed ? \\*
\textit{never}
& what is \textbf{never} the color of the sleep area of the bed ? \\*
\textit{by no means}
& what is \textbf{by no means} the color of the sleep area of the bed ? \\

\end{longtable}
\endgroup

\begin{figure}[t]
\centering
\begin{minipage}[t]{0.48\linewidth}
    \centering
    \includegraphics[width=\linewidth]{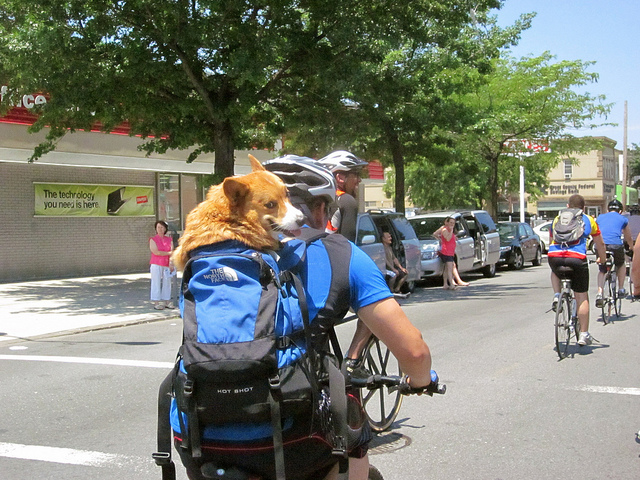}\\
    \small (a) GQA
\end{minipage}
\hfill
\begin{minipage}[t]{0.48\linewidth}
    \centering
    \includegraphics[width=\linewidth]{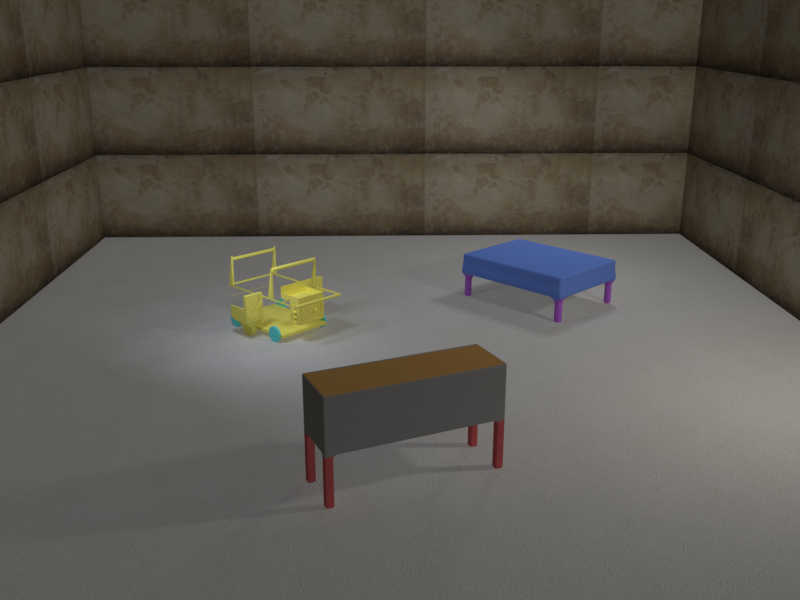}\\
    \small (b) PTR
\end{minipage}
\caption{Images accompanying the visual examples in Table~\ref{tab:benchmark_examples}. The same image is used for the original prompt and all four negated variants.}
\label{fig:benchmark_example_images}
\end{figure}

\subsection{Benchmark Validation}
\label{app:benchmark_validation}

\paragraph{Criteria.}
To verify the grammatical and semantic validity of the constructed negated questions, we validated the benchmark according to the following three criteria.
\begin{itemize}[leftmargin=1.5em]
    \item \emph{Grammaticality}: whether the negated question is grammatically well-formed.
    \item \emph{Unambiguity}: whether the negation has a single clear interpretation, without ambiguity that allows it to be interpreted in two or more ways.
    \item \emph{Negation Scope Fidelity}: whether the negation takes scope over the entire proposition of the original question, rather than locally altering its meaning in a way that changes the answer category.
\end{itemize}
We used an LLM judge (GPT-5.6 Luna) to assign a PASS/FAIL label for each of the three criteria and excluded any example that failed at least one criterion.
For each example, we provided the LLM judge with the original question, the correct answer(s) from the source benchmark, and the corresponding negated question, and instructed it to evaluate the example according to the three criteria above.
The prompt used for this validation is provided in Prompt~\ref{lst:benchmark_validation_prompt}.

RippleEdits~\citep{rippleedit} was the only source benchmark formulated as a completion task rather than a question-answering task, leaving a blank in each input and thereby making \textit{grammaticality} difficult to assess directly.
We therefore inserted a placeholder into the blank so that the input could be evaluated as a complete sentence.
The validation prompt specifically used for RippleEdits is provided in Prompt~\ref{lst:benchmark_validation_prompt_rippleedits}.

\paragraph{Validation Results.}
The validation results are summarized in Table~\ref{tab:benchmark_validation_results}. 
\textit{Grammaticality} was rarely an issue, and \textit{unambiguity} failures were relatively infrequent, accounting for 5.07\% of the examples.
In contrast, most of the filtering was associated with \textit{negation scope fidelity}, for which 16.90\% of the examples received a FAIL label.
Overall, 21,708 of the 126,756 examples (17.13\%) failed at least one criterion, leaving 105,048 examples for use in our study.
The PASS rates were consistent across the six source benchmarks, as shown in Table~\ref{tab:benchmark_validation_by_benchmark}.

Notably, among the examples that failed \textit{negation scope fidelity}, 84.29\% used \textit{never} as the negation expression.
To investigate this pattern, we analyzed the reasons generated by GPT-5.6 Luna alongside its validation judgments.
We found that these failures were primarily attributed to the temporal meaning introduced by \textit{never}: the LLM judge interpreted this additional temporal meaning as altering the semantics of the original question and therefore judged the negated question as failing to preserve the scope of the original proposition.

\begin{table}[h]
\centering
\begin{tabular}{lccc}
\toprule
Validation Criterion & PASS & FAIL & FAIL Rate (\%) \\
\midrule
Grammaticality & 125,527 & 1,229 & 0.97 \\
Unambiguity & 120,329 & 6,427 & 5.07 \\
Negation Scope Fidelity & 105,330 & 21,426 & 16.90 \\
\midrule
\textbf{Overall} & \textbf{105,048} & \textbf{21,708} & \textbf{17.13} \\
\bottomrule
\end{tabular}
\caption{Benchmark validation results for the negation benchmark dataset. Overall rejection indicates that a record was rejected if it failed at least one validation criterion.}
\label{tab:benchmark_validation_results}
\end{table}

\begin{table}[h]
\centering
\begin{tabular}{lccc}
\toprule
Benchmark & Original & PASS & PASS Rate (\%) \\
\midrule
PopQA          & 45,476 & 37,426 & 82.30 \\
RippleEdits    & 18,488 & 14,150 & 76.54 \\
PhantomWiki    & 6,660  & 5,668  & 85.11 \\
SynthWorlds & 1,120  & 840    & 75.00 \\
GQA            & 21,112 & 15,912 & 75.37 \\
PTR            & 33,900 & 31,052 & 91.60 \\
\midrule
\textbf{Overall} & \textbf{126,756} & \textbf{105,048} & \textbf{82.87} \\
\bottomrule
\end{tabular}
\caption{Benchmark validation results by source benchmark. Original denotes the number of examples in the original benchmark, while PASS denotes the number of examples that passed our filtering process. We also report the corresponding pass rate.}
\label{tab:benchmark_validation_by_benchmark}
\end{table}

\paragraph{Human Validation.}
To assess the reliability of the LLM-based filtering, we conducted human validation.
Three authors independently annotated 240 examples using the same three criteria as the LLM judge.
Specifically, we randomly sampled 10 examples from each of the six source benchmarks, maintaining a 4:1:5 ratio across the train, validation, and test splits, respectively.
Each sampled example was paired with all four negation expressions, resulting in \(6 \times 10 \times 4 = 240\) examples in total.

Human--human agreement results are reported in Table~\ref{tab:human_human_agreement_benchmark_validation}.
Unanimous agreement was very high for \textit{grammaticality} and \textit{unambiguity}, exceeding 98\% for both criteria.
For \textit{negation scope fidelity}, the unanimous agreement rate was lower at 77.08\%, largely due to disagreement on examples using \textit{never}, for which the unanimous agreement rate was 71.11\%.
Across the three human annotators, Fleiss' $\kappa$ was 0.440.
This value is attributable in part to the highly skewed label distribution toward PASS, indicating that the vast majority of examples were judged to be valid under the validation criteria.

Human--LLM agreement results are reported in Table~\ref{tab:human_llm_agreement_benchmark_validation}.
We first determined a human majority label for each example and criterion based on agreement by at least two of the three annotators.
We then compared this label with the corresponding LLM judgment. Agreement was high for \textit{grammaticality} and \textit{unambiguity}, exceeding 90\% for both criteria, while \textit{negation scope fidelity} showed a slightly lower agreement rate of 84.58\%.
Cohen's $\kappa$ was 0.334, which is again partly attributable to the label distribution being heavily skewed toward PASS.
As discussed above, GPT-5.6 Luna applied a particularly strict interpretation of \textit{negation scope fidelity}, especially for examples involving \textit{never}.
This suggests that the LLM-based filtering was conservative with respect to examples that were difficult to judge.

\begin{table}[h]
\centering
\begin{tabular}{lccc}
\toprule
Validation Criterion & \makecell{Unanimous \\ PASS} & \makecell{Unanimous \\ FAIL} & \makecell{Unanimous \\ Agreement (\%)} \\
\midrule
Grammaticality & 237 (98.75\%) & 2 (0.83\%) & 99.58\% \\
Unambiguity & 237 (98.75\%) & 0 (0.00\%) & 98.75\% \\
Negation Scope Fidelity & 174 (72.50\%) & 11 (4.58\%) & 77.08\% \\
\midrule
\textbf{Overall} & \textbf{648 (90.00\%)} & \textbf{13 (1.81\%)} & \textbf{91.81\%} \\
\bottomrule
\end{tabular}
\caption{Human--human agreement for benchmark validation by validation criterion. Percentages are calculated over all criterion-level decisions within each validation criterion.}
\label{tab:human_human_agreement_benchmark_validation}
\end{table}

\begin{table}[h]
\centering
\begin{tabular}{lccc}
\toprule
Validation Criterion & \makecell{Agreed \\ PASS} & \makecell{Agreed \\ FAIL}
& \makecell{Human--LLM \\ Agreement (\%)} \\
\midrule
Grammaticality & 230 (95.83\%) & 2 (0.83\%) & 96.67\% \\
Unambiguity & 217 (90.42\%) & 0 (0.00\%) & 90.42\% \\
Negation Scope Fidelity & 185 (77.08\%) & 18 (7.50\%) & 84.58\% \\
\midrule
\textbf{Overall} & \textbf{632 (87.78\%)} & \textbf{20 (2.78\%)} & \textbf{90.56\%} \\
\bottomrule
\end{tabular}
\caption{Human--LLM agreement for benchmark validation by validation criterion. Agreed PASS and Agreed FAIL indicate decisions for which the human majority label and the LLM label were identical. Percentages are calculated over all decisions within each criterion.}
\label{tab:human_llm_agreement_benchmark_validation}
\end{table}

\begin{lstlisting}[caption={Template for GPT-5.6 Luna benchmark validation.},
label={lst:benchmark_validation_prompt}]
You are an expert dataset validator. Evaluate the quality of a negation QA dataset.

Each example contains:

- an Affirmative question,
- Correct answer(s) from an existing QA dataset, and
- a Negated question derived from the Affirmative question.

The dataset tests whether an LLM can answer the Negated question with an answer that does not satisfy the original affirmative relation.

Evaluate exactly three independent properties.

1. NEGATED_QUESTION_GRAMMATICALITY

Determine whether the Negated question is grammatically well-formed and interpretable.
Judge grammaticality, not naturalness or pragmatic plausibility. Do not fail a question merely because it is unusual in ordinary conversation.
PASS if the question is grammatically well-formed and interpretable; otherwise FAIL.

2. NEGATION_UNAMBIGUITY

Determine whether the scope and target of the negation are clear.
Do not confuse multiple valid answers with ambiguity. A Negated question may have many valid answers; this alone is NOT a defect.
PASS if it is clear what relation, property, or condition is being negated; otherwise FAIL.

3. NEGATION_SCOPE_FIDELITY

Determine whether the negation takes scope over the entire proposition expressed by the Affirmative question. Here, the proposition means the complete statement about a candidate answer that would make it a valid answer to the Affirmative question.
The negation should apply to this proposition as a whole, rather than negating only part of its semantic content, such as a subject, relation, property, or condition within the proposition.
PASS if the Negated question negates the original proposition while otherwise preserving its semantic content; otherwise FAIL.

Important rules:

- Evaluate semantic meaning rather than exact string matching.
- Do not require the Negated question to have a unique answer.
- Do not penalize unusual but grammatical Negated questions.
- Distinguish grammaticality from naturalness.
- Distinguish semantic ambiguity from having multiple valid answers.
- Evaluate all three criteria independently.

Return only the JSON fields required by the supplied response schema.

Now evaluate:

Affirmative question:
{{AFFIRMATIVE_QUESTION}}

Correct answer(s):
{{CORRECT_ANSWERS}}

Negated question:
{{NEGATED_QUESTION}}

\end{lstlisting}

\begin{lstlisting}[caption={Template for GPT-5.6 Luna benchmark validation (used specifically when the source benchmark is RippleEdits).},
label={lst:benchmark_validation_prompt_rippleedits}]
You are an expert dataset validator. Evaluate the quality of a negation sentence-completion dataset.

Each example contains:

- an Affirmative completion template,
- Correct answer(s) from an existing sentence-completion dataset,
- a Negated completion template derived from the Affirmative completion template.

The token [CANDIDATE_ANSWER] is a placeholder for a grammatically compatible noun phrase of the expected answer type. Treat each template as a complete sentence obtained by replacing [CANDIDATE_ANSWER] with such a phrase. Evaluate the completed sentence template, not the raw prefix before the placeholder was inserted.

The dataset tests whether an LLM can complete the Negated completion template with an answer that does not satisfy the original affirmative relation.

Evaluate exactly three independent properties.

1. NEGATED_COMPLETION_GRAMMATICALITY

Determine whether the Negated completion template forms a grammatically well-formed and interpretable sentence when [CANDIDATE_ANSWER] is replaced by a grammatically compatible answer phrase.
Judge grammaticality, not naturalness or pragmatic plausibility. Do not fail a completion merely because it is unusual in ordinary conversation or because the placeholder is abstract.
PASS if the completed sentence template is grammatically well-formed and interpretable; otherwise FAIL.

2. NEGATION_UNAMBIGUITY

Determine whether the scope and target of the negation are clear.
Do not confuse multiple valid completions with ambiguity. A Negated completion task may have many valid answers; this alone is NOT a defect.
PASS if it is clear what relation, property, or condition is being negated; otherwise FAIL.

3. NEGATION_SCOPE_FIDELITY

Determine whether the Negated completion template negates the entire proposition expressed by the Affirmative completion template. Here, the proposition means the complete statement about [CANDIDATE_ANSWER] that would make it a valid completion of the Affirmative template.
The negation should apply to this proposition as a whole, rather than negating only part of its semantic content, such as a subject, relation, property, or condition within the proposition.
PASS if the Negated completion template negates the original proposition while otherwise preserving its semantic content; otherwise FAIL.

Important rules:

- Evaluate semantic meaning rather than exact string matching.
- Treat [CANDIDATE_ANSWER] as a syntactically and semantically compatible answer phrase.
- Do not require the Negated completion task to have a unique answer.
- Do not penalize unusual but grammatical Negated completion templates.
- Distinguish grammaticality from naturalness.
- Distinguish semantic ambiguity from having multiple valid answers.
- Evaluate all three criteria independently.

Return only the JSON fields required by the supplied response schema.

Now evaluate:

Affirmative question:
{{AFFIRMATIVE_QUESTION}}

Correct answer(s):
{{CORRECT_ANSWERS}}

Negated question:
{{NEGATED_QUESTION}}

\end{lstlisting}

\section{Evaluation Details}
\label{app:evaluation_details}

\paragraph{Criteria.}
To assess how well a model's answers conform to the intended behavior under negation, we evaluated each model prediction according to the following three criteria.
\begin{itemize}[leftmargin=1.5em]
    \item \emph{Original answer correctness}: whether the model's answer to the original question is semantically equivalent to at least one of the correct reference answer(s) provided by the source benchmark.
    \item \emph{Repetition under negation}: whether the model's answer to the negated question is semantically equivalent to its own answer to the original question -- that is, whether the model echoed its original answer instead of engaging with the negation.
    \item \emph{Negated answer category validity}: whether the model's answer to the negated question is a contextually and semantically appropriate kind of answer for that question, irrespective of whether it is factually correct for the particular subject.
\end{itemize}
We used an LLM judge (GPT-5.6 Luna) to assign a PASS/FAIL/UNCERTAIN label to each criterion, and instructed it to ignore factual correctness when judging the third criterion, since a well-formed but factually wrong answer should still be scored as a valid alternative. The prompts used for this evaluation are provided in Prompts~\ref{lst:negation_evaluation_prompt_property1}--\ref{lst:negation_evaluation_prompt_property3}.

\paragraph{Evaluation Procedure.}
We evaluate each criterion with a separate, isolated LLM-judge call rather than asking the judge to assess all three criteria at once.
Because the first two criteria gate the third by construction (a negated answer can only be scored for repetition if the original answer was itself correct, and can only be scored for answer category validity if it did not simply repeat the original answer), we apply the three judge calls as a cascade: we evaluate original answer correctness on every pair; among pairs that pass, we evaluate repetition under negation; and among pairs whose negated answer does \emph{not} repeat the original answer, we evaluate negated answer category validity. Pairs judged UNCERTAIN on a criterion, or that do not reach a later stage of the cascade, are excluded from the corresponding downstream criterion. We use greedy decoding with vLLM to obtain predictions from all open-source models. Aggregate results across models are reported in \S~\ref{sec:evaluation}.

\paragraph{Human Validation.}
To assess the reliability of the LLM-based evaluation, we conducted human validation using a newly sampled set of 240 examples, following the same sampling procedure described in Appendix~\ref{app:benchmark_validation} (six source benchmarks $\times$ ten examples $\times$ four negation forms, maintaining the 4:1:5 train/validation/test ratio).
Three authors independently annotated \textbf{gemma-3-12b-it} for the same three criteria, and we compare their judgments against the LLM judge's labels.

Human--human agreement results are reported in Table~\ref{tab:human_human_agreement_negation_evaluation}. Unanimous agreement across all three annotators was high for all three criteria (84.17\%--94.58\% per criterion), and Fleiss' $\kappa$ across the three annotators was 0.8783.

Human--LLM agreement results are reported in Table~\ref{tab:human_llm_agreement_negation_evaluation}. As with the benchmark-validation annotation, we determined a human majority label for each example and criterion (agreement by at least two of the three annotators) and compared it against the LLM judge's label. Because repetition under negation and answer category validity are only defined for the subset of examples that reach that stage of the cascade (\S~\ref{app:evaluation_details}, \emph{Evaluation Procedure}), examples outside a criterion's applicable subset contribute a null decision for that criterion and are excluded from its agreement calculation; percentages in Table~\ref{tab:human_llm_agreement_negation_evaluation} are computed over these non-null human--LLM decision pairs only. Agreement was high across all three criteria (84.62\%--98.04\% per criterion, 96.06\% overall), and Cohen's $\kappa$ was 0.9212. Together, these results indicate that the LLM judge's evaluation of model answers under negation closely tracks human judgment.

\begin{lstlisting}[caption={Template for GPT-5.6 Luna negation evaluation (Property 1: affirmative answer correctness).},
label={lst:negation_evaluation_prompt_property1}]
You are an expert answer evaluator. Evaluate the correctness of a model's answers on a negation QA benchmark.

Each example contains:
- an Affirmative question and the model's answer to it,
- a Negated question (derived from the Affirmative question) and the model's answer to it, and
- Correct answer(s) for the Affirmative question.

The dataset tests whether an LLM can answer the Negated question with an answer that does not satisfy the original affirmative relation.

Evaluate exactly one property.

1. POSITIVE_PREDICTION_MATCHES_CORRECT_ANSWER
Determine whether the model's answer to the Affirmative question is semantically equivalent to one of the Correct answer(s). When multiple Correct answers are listed, they are alternative acceptable phrasings of the same answer (e.g. synonyms or aliases), not a set the model must name in full: it is sufficient for the model's answer to be semantically equivalent to any single one of them. Answer PASS if it is equivalent to at least one; otherwise answer FAIL.

Important rules:
- Evaluate semantic meaning rather than exact string matching.
- Do not infer or return an overall success label.

Return only the JSON fields required by the supplied response schema.

Now evaluate:

Affirmative question:
{{AFFIRMATIVE_QUESTION}}

Affirmative model answer:
{{AFFIRMATIVE_ANSWER}}

Correct answer(s):
{{CORRECT_ANSWERS}}

\end{lstlisting}

\begin{lstlisting}[caption={Template for GPT-5.6 Luna negation evaluation (Property 2: repetition under negation).},
label={lst:negation_evaluation_prompt_property2}]
You are an expert answer evaluator. Evaluate the correctness of a model's answers on a negation QA benchmark.

Each example contains:
- an Affirmative question and the model's answer to it,
- a Negated question (derived from the Affirmative question) and the model's answer to it, and
- Correct answer(s) for the Affirmative question.

The dataset tests whether an LLM can answer the Negated question with an answer that does not satisfy the original affirmative relation.

Evaluate exactly one property.

2. NEGATED_PREDICTION_MATCHES_CORRECT_ANSWER
Determine whether the model's answer to the Negated question is semantically equivalent to its answer to the Affirmative question -- that is, whether the model gave the same answer to the Negated question that it originally gave to the Affirmative question. A different surface form that refers to the same thing (a synonym, a reordering, an equivalent phrase) still counts as the same answer; do not confuse a merely different phrasing with a genuinely different answer. Answer PASS if the two answers are the same (the model, in effect, echoed its original answer instead of engaging with the negation); otherwise answer FAIL.

Important rules:
- Evaluate semantic meaning rather than exact string matching.
- Do not infer or return an overall success label.

Return only the JSON fields required by the supplied response schema.

Now evaluate:

Affirmative question:
{{AFFIRMATIVE_QUESTION}}

Affirmative model answer:
{{AFFIRMATIVE_ANSWER}}

Negated question:
{{NEGATED_QUESTION}}

Negated model answer:
{{NEGATED_ANSWER}}

\end{lstlisting}

\begin{lstlisting}[caption={Template for GPT-5.6 Luna negation evaluation (Property 3: negated answer category validity).},
label={lst:negation_evaluation_prompt_property3}]
You are an expert answer evaluator. Evaluate the correctness of a model's answers on a negation QA benchmark.

Each example contains:
- an Affirmative question and the model's answer to it,
- a Negated question (derived from the Affirmative question) and the model's answer to it, and
- Correct answer(s) for the Affirmative question.

The dataset tests whether an LLM can answer the Negated question with an answer that does not satisfy the original affirmative relation.

Evaluate exactly one property.

3. NEGATED_PREDICTION_HAS_EXPECTED_ANSWER_TYPE
Determine whether the model's answer to the Negated question is a contextually and semantically appropriate kind of answer, given the question's intended meaning as a whole -- not just judged against part of it (e.g. matching only the category named in the question while ignoring the negation attached to it). Ignore whether the answer is factually correct for the particular subject. A well-formed answer of the right kind is PASS even when it is factually wrong; an answer of the wrong kind (empty, a non-answer, or a different category of thing entirely) is FAIL.

Important rules:
- Evaluate semantic meaning rather than exact string matching.
- For property 3, factual correctness is irrelevant: judge only the shape/kind of the answer, never whether it is true.
- Do not infer or return an overall success label.

Return only the JSON fields required by the supplied response schema.

Now evaluate:

Negated question:
{{NEGATED_QUESTION}}

Negated model answer:
{{NEGATED_ANSWER}}

\end{lstlisting}

\begin{table}[h]
\centering
\begin{tabular}{lccc}
\toprule
Evaluation Criterion & \makecell{Unanimous \\ PASS} & \makecell{Unanimous \\
FAIL} & \makecell{Unanimous \\ Agreement (\%)} \\
\midrule
\makecell[l]{Original Answer Correctness} & 91 (37.92\%) & 135 (56.25\%) & 94.17\% \\
\makecell[l]{Repetition Under Negation} & 95 (39.58\%) & 132 (55.00\%) & 94.58\% \\
\makecell[l]{Negated Answer Category Validity} & 179 (74.58\%) & 23 (9.58\%) & 84.17\% \\
\midrule
\textbf{Overall} & \textbf{365 (50.69\%)} & \textbf{290 (40.28\%)} & \textbf{90.97\%} \\
\bottomrule
\end{tabular}
\caption{Human--human agreement for negation evaluation by evaluation criterion. Percentages are calculated over all criterion-level decisions within each evaluation criterion.}
\label{tab:human_human_agreement_negation_evaluation}
\end{table}

\begin{table}[h]
\centering
\begin{tabular}{lccc}
\toprule
Evaluation Criterion & \makecell{Agreed \\ PASS} & \makecell{Agreed \\ FAIL} & \makecell{Human--LLM \\ Agreement (\%)} \\
\midrule
\makecell[l]{Original Answer Correctness} & 96 (40.00\%) & 137 (57.08\%) & 97.08\% \\
\makecell[l]{Repetition Under Negation} & 62 (60.78\%) & 38 (37.25\%) & 98.04\% \\
\makecell[l]{Negated Answer Category Validity} & 33 (84.62\%) & 0 (0.00\%) & 84.62\% \\
\midrule
\textbf{Overall} & \textbf{191 (50.13\%)} & \textbf{175 (45.93\%)} & \textbf{96.06\%} \\
\bottomrule
\end{tabular}
\caption{Human--LLM agreement for negation evaluation by evaluation criterion. Agreed PASS and Agreed FAIL indicate decisions for which the human majority label and the LLM label were identical. Percentages are calculated over non-null human--LLM decision pairs
within each criterion.}
\label{tab:human_llm_agreement_negation_evaluation}
\end{table}

\clearpage
\section{Experiment Details}
\label{app:experiment_details}

\subsection{Details on the Information Flow Experiment}
\label{app:information_flow}

We describe the activation patching used in
\S~\ref{sec:information_flow}. To trace how information from
negation tokens reaches the final prompt token, we use the
token groups defined in \S~\ref{sec:negation-localization}.

\paragraph{Attention contributions.}
To isolate the information flow from a token group $G$ to a
destination position $t$, we sum the attention-weighted value
vectors from that group across all attention heads:
\[
\mathbf c^\ell_{G\to t}(x)
=
\sum_h \mathbf W_O^{\ell,h}
\left(
\sum_{s\in G(x)}
\alpha^{\ell,h}_{t,s}(x)\,
\mathbf v^{\ell,h}_s(x)
\right),
\]
where $\alpha^{\ell,h}_{t,s}$ is the attention weight,
$\mathbf v^{\ell,h}_s$ is the value vector, and
$\mathbf W_O^{\ell,h}$ is the output projection matrix for
attention head $h$.
We retain the model's attention weights without renormalizing
them within the selected group. The effect of this intervention
can therefore reflect changes in attention to these tokens,
the information carried by their value vectors, or both.

\paragraph{Activation patching.}
We first compute $\mathbf c^\ell_{G\to t}$ from the activations
of the original and negated prompts.
During an intervention, we remove the contribution from the
current execution and replace it with that from the paired prompt:
\[
\widetilde{\mathbf o}^{\ell}_t
=
\mathbf o^{\ell,\mathrm{current}}_t
-
\mathbf c^{\ell,\mathrm{current}}_{G\to t}
+
\mathbf c^{\ell,\mathrm{paired}}_{G\to t'},
\]
where $\mathbf o^\ell_t$ is the attention output and $t'$ is
the corresponding position in the paired prompt.

\paragraph{Implementation in Qwen3.5.}
Qwen3.5 uses both softmax and linear attention, so we adapt the
decomposition to each layer type.
For softmax-attention layers, we use the decomposition above.
For linear-attention layers, we isolate each token group's contribution
by retaining only that group's value vectors after the local convolution,
while keeping queries, keys, and gates unchanged.
We use the normalization scale computed from the full output, rather
than normalizing each group's contribution separately.
We then replace the resulting contribution with that from the paired
prompt using the same intervention procedure.

\clearpage

\subsection{Details on the Negation Signal Experiment}
\label{app:negation_signal}

\paragraph{Negation-Induced Differences}

We compute the negation-induced differences at the final prompt
token as defined in \S~\ref{sec:negation-localization}.
At each layer, we first average these differences over training
pairs within each benchmark:
\[
\mathbf d_k^\ell
=
\frac{1}{|\mathcal D_k^{\mathrm{train}}|}
\sum_{(i,f)\in\mathcal D_k^{\mathrm{train}}}
\Delta\mathbf h_{i,f}^{\ell}.
\]
We then average the benchmark-specific means with equal weights:
\[
\mathbf d^\ell
=
\frac{1}{K}\sum_{k=1}^{K}\mathbf d_k^\ell,
\]
where $K$ is the number of benchmarks.

To assess whether the individual differences are aligned across
examples, we compute their average similarity to the mean vector. We compute training and validation similarities
separately and use only training differences to estimate
$\mathbf d^\ell$.

\paragraph{Steering}

We subtract the mean difference from the negated prompt's
final-token representation:
\[
\mathbf h_{\mathrm{int}}^\ell(x_{i,f}^{\mathrm{neg}})
=
\mathbf h^\ell(x_{i,f}^{\mathrm{neg}})
-
\beta\mathbf d^\ell,
\]
where $\beta$ controls the steering strength.
We apply the intervention after the selected transformer block,
leaving other token positions unchanged.
The intervention is applied once during prompt processing
and is not repeated at generated answer tokens.

\paragraph{Downstream Attention and MLP Outputs}

We examine whether steering changes downstream attention and
MLP outputs in the direction of their values under the original
prompt. Let $\mathbf o_c(x)$ denote the output of attention or
MLP component $c$ at the final prompt token.
We compare the intervention-induced difference
\[
\Delta\mathbf o_c^{\mathrm{int}}
=
\mathbf o_c(x_{i,f}^{\mathrm{neg}};\mathrm{intervention})
-
\mathbf o_c(x_{i,f}^{\mathrm{neg}})
\]
with the difference between the unmodified paired prompts:
\[
\Delta\mathbf o_c^{\mathrm{natural}}
=
\mathbf o_c(x_i)
-
\mathbf o_c(x_{i,f}^{\mathrm{neg}}).
\]
Both differences are measured relative to the same unmodified
negated prompt. We compute their cosine similarity separately
for attention and MLP outputs at each downstream layer.

\clearpage
\subsection{Details on the Component Identification Experiment}
\label{app:component_identification}

We describe how we select attention heads and MLP neurons,
evaluate their causal effects, and analyze their outputs.
We use the notation and evaluation metrics introduced in
\S~\ref{sec:negation-localization}.
Components are selected using training examples and fixed before
validation. We select only from layers after the layer used for
negation-signal removal (\S~\ref{sec:negation_signal}).
We use the top five heads and top 256 neurons for the main analysis.
We additionally evaluate the top one and ten heads and the top
64 and 1,024 neurons.

\paragraph{Component Identification.}

We aim to identify attention heads whose attention changes
consistently under negation and MLP neurons whose activation
changes favor the alternative.
To do so, we score each component for each paired prompt,
average scores within each benchmark, and then equally across
the six benchmarks.
We also retain benchmark- and relation-specific scores to examine
the consistency of the selected components.
We define the scores as follows.

For attention heads, we measure the attention-weight difference from
the final prompt token:
\[
S_{\ell,h}^{\mathrm{attn}}(i,f)
=
\left\|
\boldsymbol{\alpha}_{\ell,h}(x_{i,f}^{\mathrm{neg}})
-
\boldsymbol{\alpha}_{\ell,h}(x_i)
\right\|_2.
\]
To examine which tokens gain or lose attention, we also measure
attention-mass changes for the token groups defined in
\S~\ref{sec:information_flow}.

For MLP neurons, we define a neuron activation $m_{\ell,j}$ as one coordinate of
the elementwise product of the activated gate projection and
the up-projection, before the MLP down-projection.
Following attribution patching~\citep{syed-etal-2024-attribution},
we compute
\[
S_{\ell,j}^{\mathrm{MLP}}(i,f)
=
\left[
m_{\ell,j}(x_{i,f}^{\mathrm{neg}})
-
m_{\ell,j}(x_i)
\right]
\left.
\frac{\partial g_{i,f}}{\partial m_{\ell,j}}
\right|_{x_{i,f}^{\mathrm{neg}}}.
\]
The gradient is evaluated on the negated prompt using the
full-answer score gap $g_{i,f}$.
For example, if the model answers ``Spain'' for the original
prompt and ``France'' for ``Madrid is not the capital of
[BLANK],'' this gap measures preference for ``France'' over
``Spain.''
A positive attribution score estimates that restoring the
original activation would reduce this preference.
We rank neurons by signed attribution, use this approximation
to select neurons, and evaluate their actual effects through
interventions.

\paragraph{Causal Interventions.}

We replace selected head outputs or neuron activations in the
negated execution with their values from the original execution.
Head outputs are replaced before the attention output projection,
and neuron activations before the MLP down-projection.
Interventions affect only the final prompt token and are not
repeated at generated token positions.

We evaluate (1) the five selected heads individually,
(2) each selected head set and neuron set jointly, and
(3) the top five heads together with the top 256 neurons.
For each set intervention, we use five random component sets
with the same number of components in each layer.

\paragraph{Logit Lens Analysis.}

We apply a logit lens to the original component output,
the negated component output, and their difference.
Head outputs are first mapped through the corresponding attention
output projection, and neuron activations through the MLP
down-projection.

For models with post-attention and post-MLP normalization,
such as Gemma 3-IT, we include the learned normalization gains
and the normalization scales from the full attention or MLP
outputs in the corresponding unmodified executions.
For paired comparisons, both outputs use the same
final-normalization scale, taken from the original execution.

We then apply the model's unembedding matrix to obtain vocabulary
scores and record the 20 highest- and lowest-scoring vocabulary
items, along with the scores and ranks of the first tokens of
the original and alternative answers.
We report individual examples and aggregate results by benchmark,
relation, and answer category.

\clearpage

\subsection{Details on the Subspace Experiment}
\label{app:subspace_experiment}

We use Distributed Alignment Search (DAS; \citealp{DAS}) to identify
two low-dimensional subspaces introduced in
\S~\ref{sec_5_representation}: an \emph{answer category space} $C$,
where interventions change the preferred answer category, and a
\emph{candidate-preference space} $L_s$, where interventions change
the preferred answer within category $s$.

\paragraph{Subspace interventions.}
DAS uses gradient-based optimization to find a subspace in which
replacing representations produces a specified change in the model's
prediction. We first describe the intervention used to learn both spaces.

Let $\mathbf h_\ell(x)$ denote the residual-stream representation at
the final prompt token, immediately before the MLP at layer $\ell$.
For an orthonormal basis $U\in\mathbb R^{d\times r}$, we replace the
component within its subspace with that of another prompt $x'$:
\begin{equation}
\widetilde{\mathbf h}_\ell(x;x',U)
=
\mathbf h_\ell(x)
+
UU^\top
\left[\mathbf h_\ell(x')-\mathbf h_\ell(x)\right].
\label{eq:subspace-replacement}
\end{equation}
The component orthogonal to $U$ remains unchanged at the intervention
site, and the model continues its computation from the modified
representation. During training, this intervention is applied only
at the final prompt token.

\paragraph{DAS Training.}

We use the subspace intervention and gradient-based optimization of
DAS, with separate objectives for answer category and candidate
preference. 
For both objectives, we train the intervened model to increase the
probability of a target answer. Let $P_U(y\mid x;x')$ denote the
probability assigned to the full answer $y$ when the model processes
$x$ with the intervention in Eq.~\ref{eq:subspace-replacement}.
During training, we compute this probability by conditioning each
answer token on the preceding tokens of the target answer.

For the \textit{candidate-preference space}, we aim to learn a subspace where replacing representations between prompts from the same answer category changes which answer the model prefers. For example, we pair ``What is the capital of Spain?'' with ``What is the capital of France?'', answered with ``Madrid'' and ``Paris'', respectively. Replacing the component within this subspace in the first prompt with that from the second should shift the model's preference from ``Madrid'' toward ``Paris''.

To learn $L_s$, we therefore pair prompts from category $s$ that
produce different answers. Let $a(x')$ denote the model's stored
generated answer to $x'$. We minimize
\begin{equation}
\mathcal L_{L_s}
=
-\mathbb E_{(x,x')\sim\mathcal P_s}
\log P_{U_{L_s}}\!\left(a(x')\mid x;x'\right),
\label{eq:candidate-subspace-loss}
\end{equation}
where $\mathcal P_s$ contains within-category prompt pairs.
The objective increases the probability of the full generated answer,
rather than only its first token.
We distinguish answers using their normalized full text, so answers
that share a first token can still form a pair.

For the \textit{answer category space}, we aim to learn a subspace where replacing representations changes
the preferred answer category without requiring a particular answer
from that category. We therefore pair prompts from different
categories and increase the probability of answers from the category
of $x'$, rather than its particular generated answer.

For each pair, we sample two reference answers,
$\mathcal T(x,x')$, from training examples in that category.
We minimize
\begin{equation}
\mathcal L_C
=
-\mathbb E_{(x,x')}
\frac{1}{|\mathcal T(x,x')|}
\sum_{y\in\mathcal T(x,x')}
\log P_{U_C}(y\mid x;x').
\label{eq:category-subspace-loss}
\end{equation}
This objective encourages a change in answer category without
requiring the model to produce the particular answer to $x'$.

\paragraph{Optimization.}
We optimize an unconstrained matrix and obtain an orthonormal basis
through QR decomposition before each intervention.
Each training step uses one prompt and four sampled comparison
prompts. We use AdamW with a learning rate of $0.03$ and no weight decay.

The initial layer comparison uses rank $4$ and $200$ training steps.
Subsequent runs use up to $3{,}000$ steps.
We select checkpoints using a held-out subset of the training data:
normalized answer-score shift for $L_s$ and category prediction
accuracy for $C$.
Rank comparisons use the same layer.
The bases are individually orthonormal, but $C$ and $L_s$ are not
constrained to be orthogonal to each other.

\paragraph{Causal Validation}

We first evaluate $L_s$ on held-out prompt pairs from the same category.
We measure whether the intervention shifts preference from the
answer to $x$ toward the answer to $x'$.
We report the normalized answer-score shift, using the unmodified
score gaps under $x$ and $x'$ for normalization.
We also report the fraction of intervened generations that match
the stored answer to $x'$.

Then, we evaluate $C$ on held-out prompt pairs from different categories.
For each category, we use four fixed training-set reference answers
and average their full-answer probabilities.
We predict the category with the largest average and measure whether
the intervention changes this prediction toward the category of $x'$.
This evaluates category preference using a fixed set of answers,
not all possible answers in each category.

We additionally apply $C$ interventions between prompts from the
same category and measure changes in preference between their
answers. This tests whether interventions within $C$ also change
preference for a particular answer.

We compare each learned space with five random orthonormal spaces
of the same rank, using identical evaluation pairs.
We also evaluate replacement of the full residual representation
and its component orthogonal to the learned space.
Results are reported separately by answer category and averaged
with equal weight across categories.

\paragraph{Analyzing Changes under Negation}

We fix the learned bases and project the representations of matched
original and negated prompts into each space.
For $G\in\{C,L_s\}$,
\begin{equation}
\mathbf z_G(x)=U_G^\top\mathbf h_\ell(x),
\qquad
\Delta\mathbf z_{G,i,f}
=
\mathbf z_G(x_{i,f}^{\mathrm{neg}})
-
\mathbf z_G(x_i).
\label{eq:subspace-negation-change}
\end{equation}

Within $L_s$, we compare the magnitude and direction of the
negation-induced changes across examples from the same answer
category. To determine which answers a direction favors, we
separately evaluate how interventions along that direction change
answer scores.

Within $C$, we compare the projected representations to examine
how answer category information changes under negation.
We distinguish this geometric comparison from causal validation:
small projected differences alone do not establish that
answer category information remains causally effective in negated
prompts.

\clearpage

\section{Extended Results}
\label{app:additional-results}
\subsection{Extended Results on Negation Evaluation}
\label{app:additional-results-evaluation}

\begin{figure}[t]
  \centering
  \includegraphics[width=\linewidth]{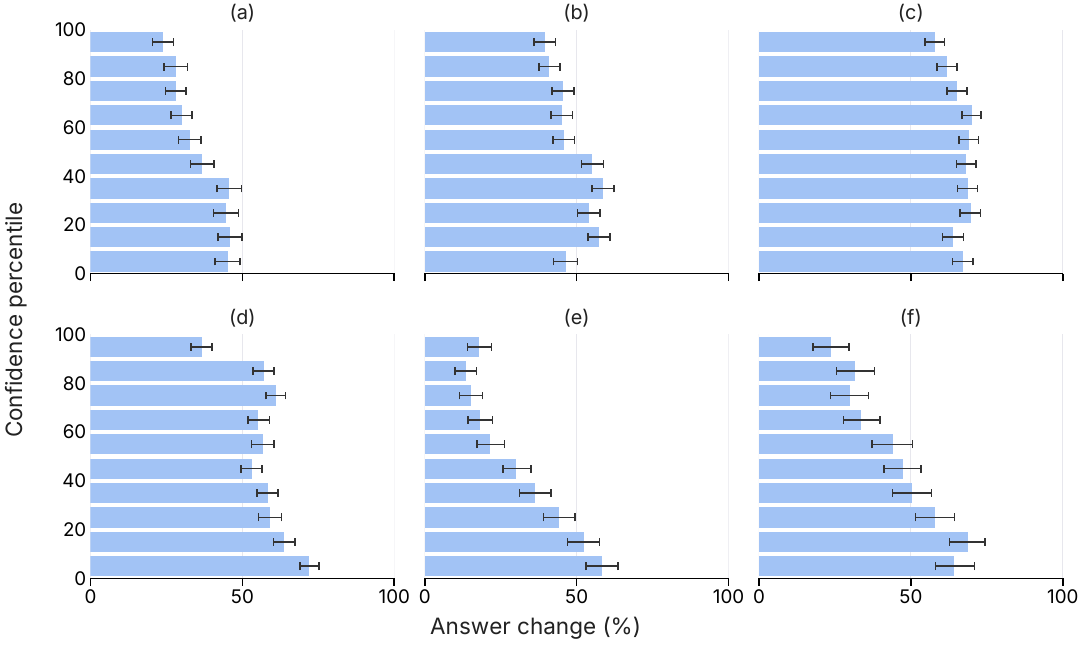}
  \caption{Answer change by confidence percentile for each model on
  the test split. Confidence is measured by mean token log-probability
  of the original answer, with percentiles computed within each model.
  We include only originally correct cases and average negation forms
  within each question. Error bars show 95\% question-bootstrap intervals.}
  \label{fig:confidence-by-model}
\end{figure}

\paragraph{Confidence and answer change by model.}
Figure~\ref{fig:confidence-by-model} breaks down the main-text result
by model. The strength and shape of the association vary across models,
rather than showing a uniform decrease in answer change.

\begin{figure}[t]
  \centering
  \includegraphics[width=\linewidth]{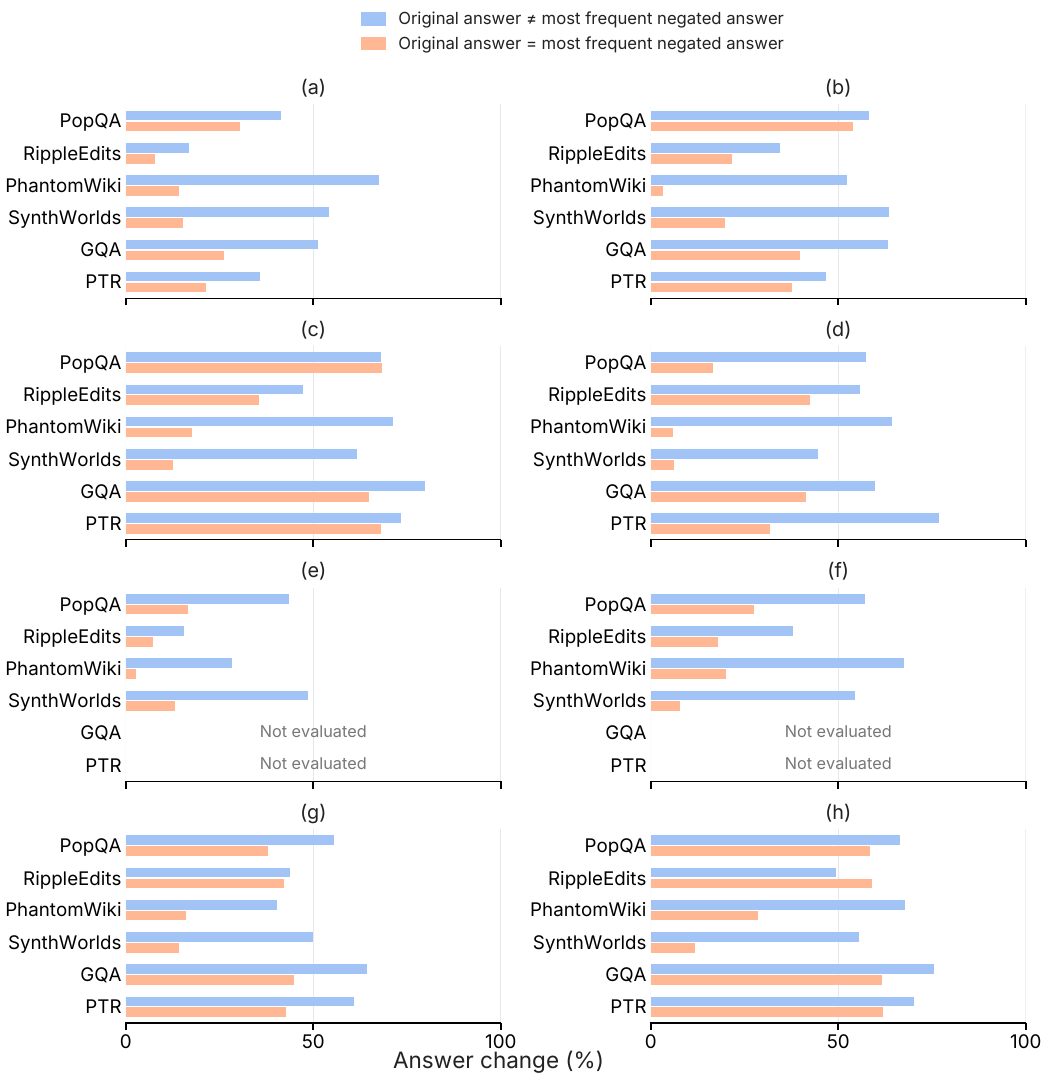}
  \caption{Answer change by model and benchmark on the test split. The most frequent negated answer is identified within each model's relation/type group using its test outputs. Rates are computed over originally correct pairs with a judged repetition outcome. Not evaluated indicates unavailable model--benchmark results.}
  \label{fig:answer-bias-by-benchmark}
\end{figure}

\paragraph{Negation bias by model and benchmark.}
Figure~\ref{fig:answer-bias-by-benchmark} separates the negation bias
comparison by model and benchmark. The matching group has a lower
answer-change rate in 42 of 44 evaluated combinations, but this
pattern is not universal.

\clearpage
\subsection{Extended Results on the Information Flow Experiment}
\label{app:information_flow}

\begin{figure*}[t]\centering
\includegraphics[width=0.9\textwidth]{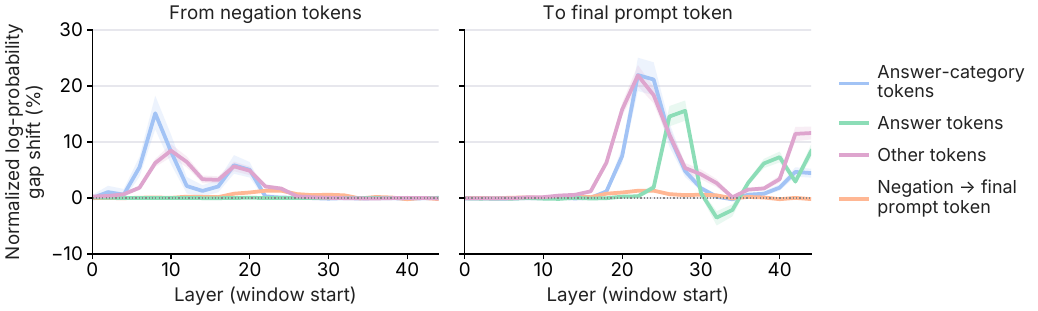}
\caption{Information flow in Gemma 3-12B-IT. \textbf{Left:} from negation tokens to other token groups and \textbf{Right:} from each group to the final prompt token.}
\label{fig:gemma4-information-flow}
\end{figure*}

Figure~\ref{fig:gemma4-information-flow} reports interventions on all
1,540 eligible Gemma 3-12B-IT validation pairs, with equal weight
across six benchmarks. Replacing contributions from negation tokens
to answer category tokens has its largest effect at L8--11,
with a normalized log-probability gap shift of 15.1\%.
Paths from answer category tokens and other tokens to the final
prompt token peak later, at L22--25, with shifts of 21.9\% each.
The direct negation-to-final-token path reaches only 1.3\%, while
answer-token contributions peak at L28--31 (15.6\%).
Layer indices are zero-based, and the horizontal axis gives the start
of each four-layer intervention window.

These peaks support an indirect route through intervening prompt
tokens, rather than a dominant direct contribution from negation
tokens. They do not identify a unique sequential circuit: each curve
comes from a separate intervention, and paths can interact downstream.
The answer-token curve also combines attention-weight and value changes;
it does not by itself distinguish reduced retrieval from a change in
the retrieved information.

\clearpage

\begin{figure}[p]
  \centering
  \includegraphics[width=\linewidth]{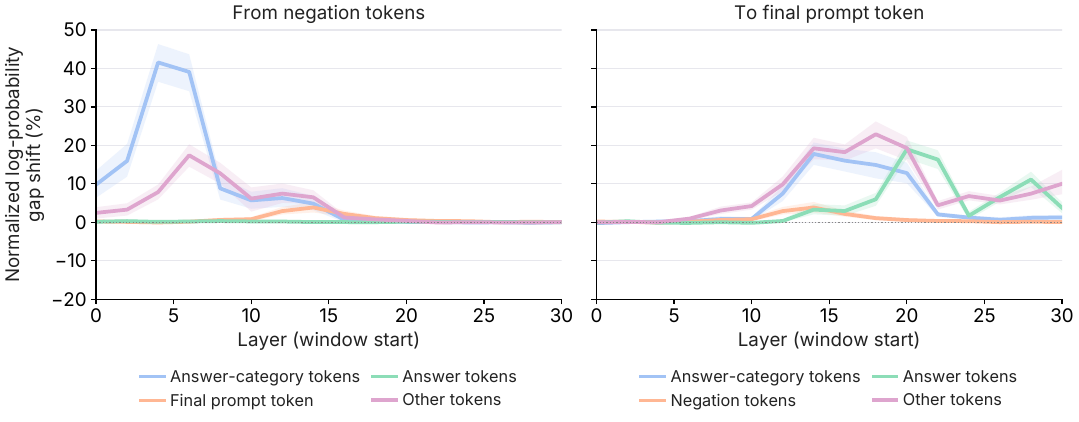}
  \vspace{0.5em}
  \includegraphics[width=\linewidth]{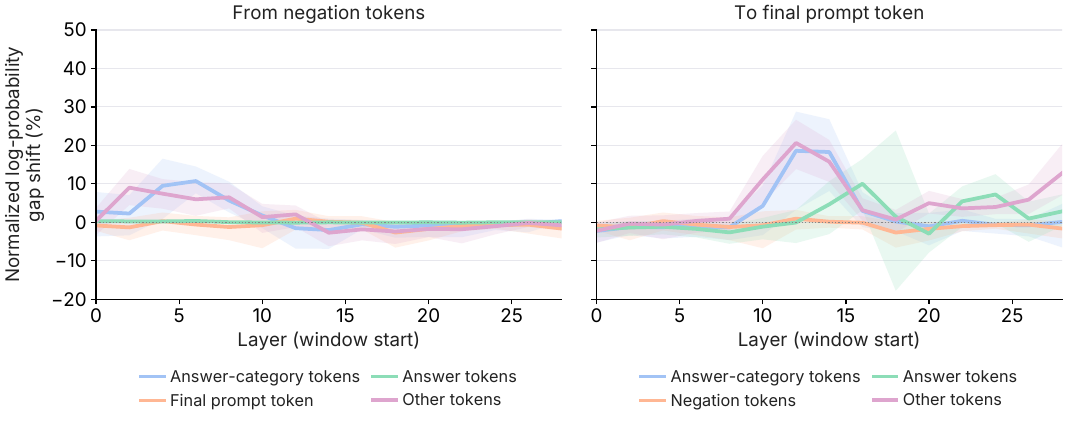}
  \caption{Information flow in Gemma 3-4B-IT (top) and
  Llama 3.1-8B-Instruct (bottom). Left: paths from negation tokens.
  Right: paths to the final prompt token. Curves show normalized
  log-probability gap shift; shading shows 95\% question-bootstrap intervals.}
  \label{fig:app-other-model-flow}
\end{figure}

\paragraph{Other models.}
Figure~\ref{fig:app-other-model-flow} extends the prompt-only analysis
to Gemma 3-4B-IT and Llama 3.1-8B-Instruct. The four-layer windows
and stride of two are unchanged, but the eligible validation cohorts
contain 925 and 423 pairs, respectively. Gemma results equally weight
six benchmarks; Llama results equally weight the four text benchmarks.
These are model-specific success cohorts, not a matched-case comparison
of model performance.

Qwen3.5-9B is evaluated separately because it combines softmax and
linear attention. Softmax layers use attention-weighted value
contributions, while linear-attention layers use token-group
contributions to memory writes after the local convolution, as described
in Appendix~\ref{app:information_flow}.
The completed analysis covers all 2,304 eligible validation pairs and
240 conditions per pair. For paths ending at the prediction position,
interventions continue during generation; other paths retain their
aligned prompt positions. Its curves therefore do not provide a
protocol-matched effect-size comparison with the prompt-only results.

\begin{figure}[t]
  \centering
  \includegraphics[width=\linewidth]{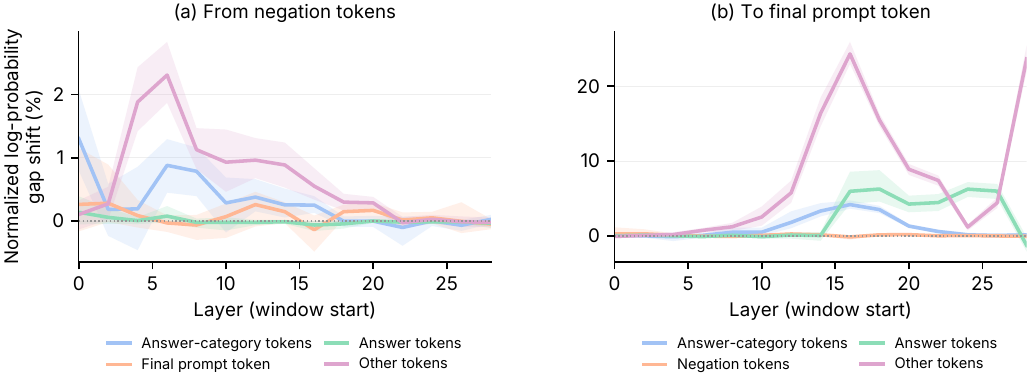}
  \caption{Information flow in Qwen3.5-9B, including softmax and linear
  attention. Final-position interventions continue during generation.
  Effects equally weight six benchmarks; shading shows 95\%
  question-bootstrap intervals.}
  \label{fig:app-qwen-flow}
\end{figure}

The unmodified Qwen execution reproduces 2,251 of 2,304 stored original
answers and 2,187 stored negated answers. The first 600 completed GQA
pairs retain their original execution settings; the remaining pairs
use prefix caching with newly computed baselines. Each intervention
is compared with the baseline from its own execution setting, and
baseline mismatches do not change the frozen cohort.


\clearpage

\subsection{Extended Results on the Negation Signal Experiment}
\label{app:negation_signal_results}

\begin{figure}[p]
  \centering
  \includegraphics[width=\linewidth]{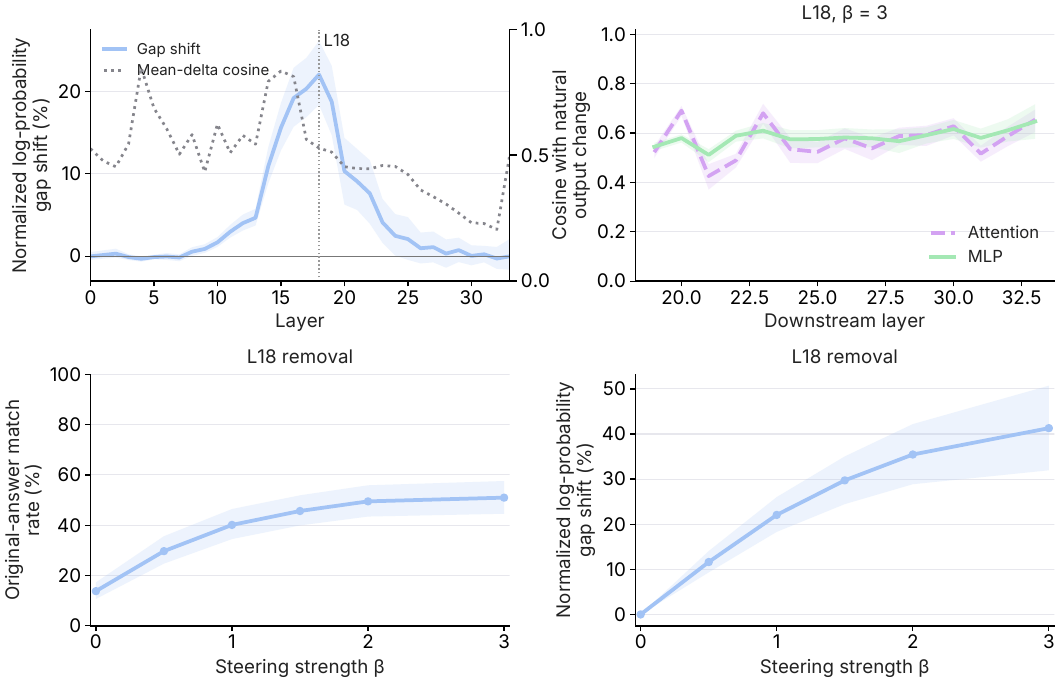}
  \vspace{0.5em}
  \includegraphics[width=\linewidth]{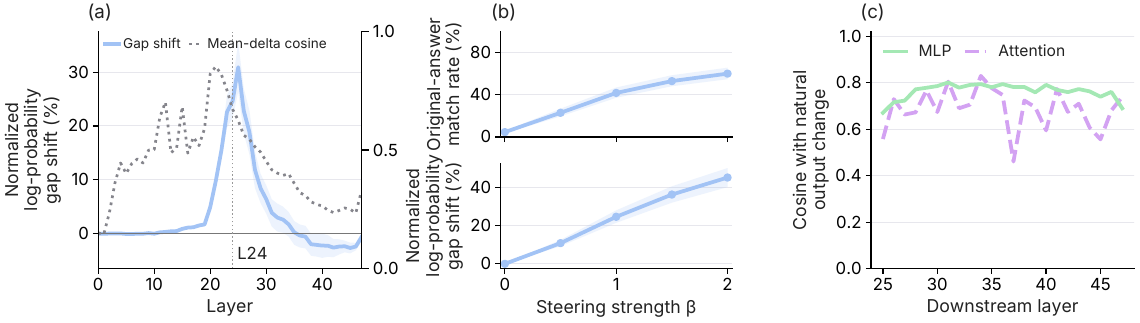}
  \caption{Negation-direction removal in Gemma 3-4B-IT (top) and
  Gemma 3-12B-IT (bottom). Dose layers are L18 and L24; downstream
  alignment uses $\beta=3$ and $\beta=2$, respectively.
  Effects equally weight six benchmarks. Shading shows available
  95\% question-bootstrap intervals; Gemma 12B downstream cosine
  is shown without a pooled interval.}
  \label{fig:app-gemma-signal}
\end{figure}

\begin{figure}[p]
  \centering
  \includegraphics[width=\linewidth]{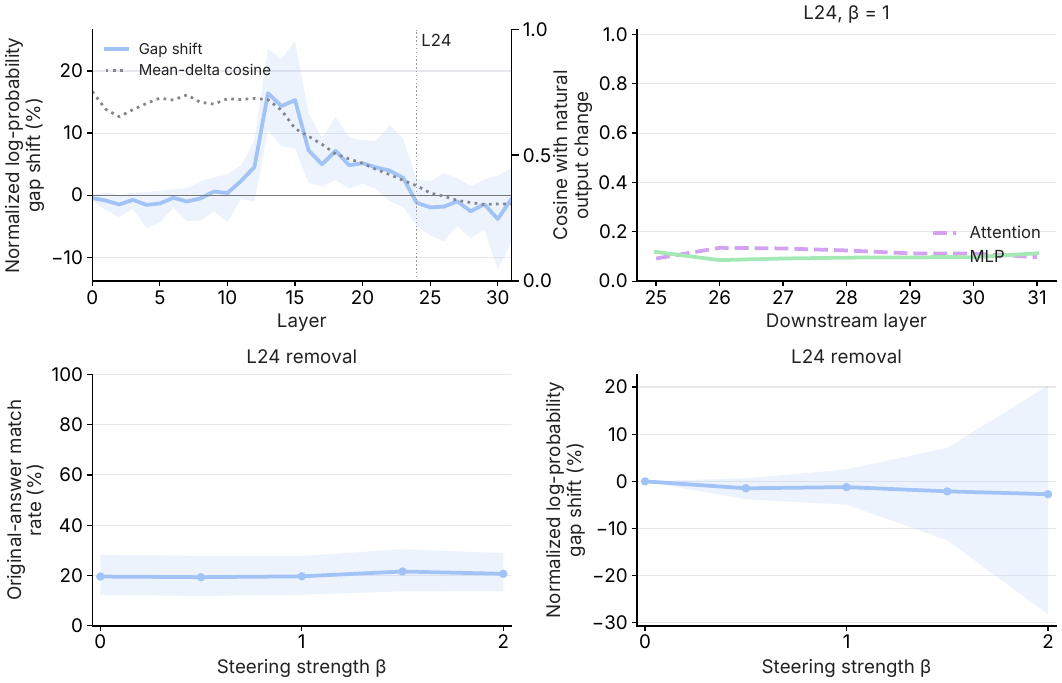}
  \vspace{0.5em}
  \includegraphics[width=\linewidth]{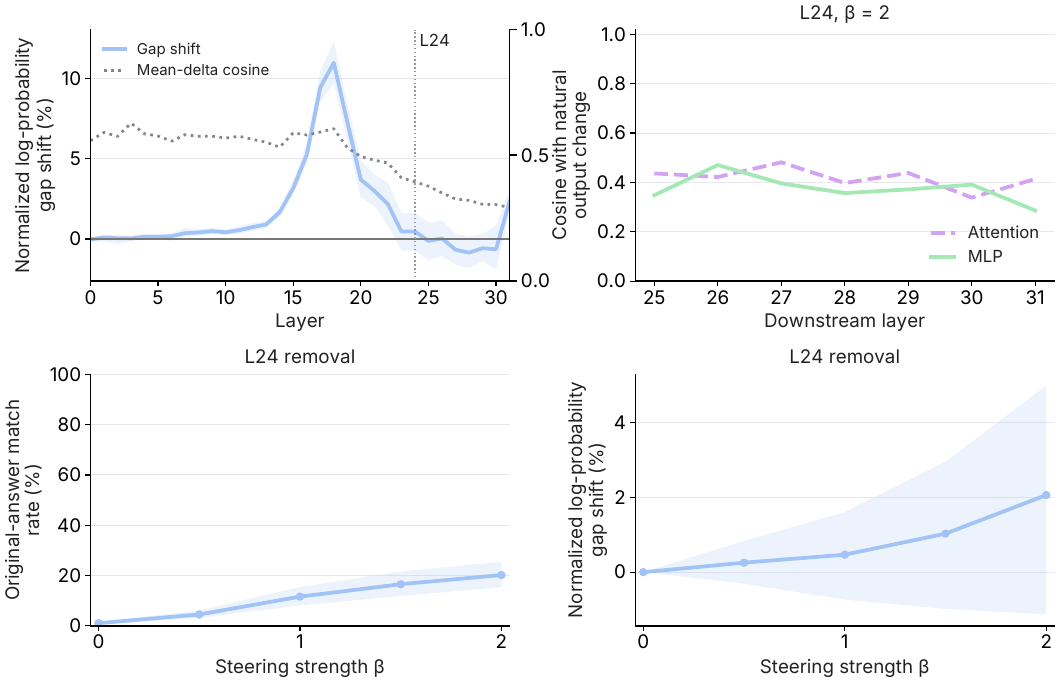}
  \caption{Negation-direction removal in Llama 3.1-8B-Instruct
  (top; four text benchmarks) and Qwen3.5-9B (bottom; six benchmarks).
  Both displayed dose experiments use L24. Downstream alignment uses
  $\beta=1$ and $\beta=2$, respectively.}
  \label{fig:app-llama-qwen-signal}
\end{figure}

We evaluate negation-direction removal in Gemma 3-4B-IT,
Gemma 3-12B-IT, Llama 3.1-8B-Instruct, and Qwen3.5-9B.
Table~\ref{tab:app-model-signal} reports the fixed layers used in the
dose experiments; Figures~\ref{fig:app-gemma-signal}
and~\ref{fig:app-llama-qwen-signal} show the layer sweeps, dose responses,
and downstream output comparisons. All interventions in these figures
affect only the final prompt token. Layer indices are zero-based.

\begin{table}[t]
\centering
\small
\setlength{\tabcolsep}{4pt}
\renewcommand{\arraystretch}{1.12}
\begin{tabular}{lrrrcc}
\toprule
Model & Pairs & Layer & \makecell{Match\\$\beta=0$} & \makecell{Match\\$\beta=1$} & \makecell{Gap shift\\$\beta=1$} \\
\midrule
Gemma 3-4B-IT & 925 & L18 & 13.8 & 40.1 [34.5, 46.4] & 22.1 [18.3, 26.0] \\
Gemma 3-12B-IT & 1540 & L24 & 4.6 & 41.8 [37.2, 46.9] & 24.6 [21.5, 28.0] \\
Llama 3.1-8B-Instruct & 423 & L24 & 19.6 & 19.7 [12.3, 27.8] & -1.2 [-4.9, 2.5] \\
Qwen3.5-9B & 2304 & L24 & 1.1 & 11.6 [8.2, 15.4] & 0.5 [-0.7, 1.6] \\
\bottomrule
\end{tabular}
\caption{Prompt-only negation-direction removal. Match denotes original-answer match rate; gap shift denotes normalized log-probability gap shift, both in percent. Brackets are 95\% question-bootstrap intervals. Results equally weight six benchmarks for Gemma and Qwen, and four text benchmarks for Llama. The fixed dose layers are not uniformly the layerwise maxima; see the full sweeps.}
\label{tab:app-model-signal}
\end{table}

The most effective layer differs across models. In the exploratory
$\beta=1$ sweep, normalized log-probability gap shift peaks at L18
for Gemma 3-4B-IT, L25 for Gemma 3-12B-IT, L13 for
Llama 3.1-8B-Instruct, and L18 for Qwen3.5-9B. 

\begin{figure}[t]
  \centering
  \includegraphics[width=\linewidth]{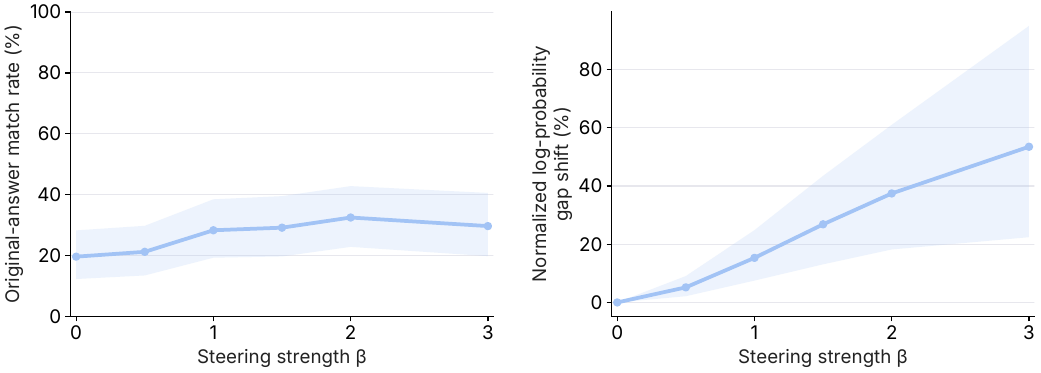}
  \caption{Prompt-only negation-direction removal at L15 in
  Llama 3.1-8B-Instruct.}
  \label{fig:app-llama-middle-dose}
\end{figure}

\clearpage
\subsection{Analysis on the Original Answer Heads}
\label{app:additional-results-components_1}

We first test whether the selected attention heads (L31H12, L25H8,
L31H13, L27H5, and L34H13) contribute to suppressing the original
answer. We evaluate Gemma 3-12B-IT on 1,540 held-out successful
validation pairs, restoring the selected outputs to their non-negated
values at the final prompt token.

For L31H13, restoring only the contribution from answer tokens
increases the original answer's full-sequence log-probability by
$0.129$ nats (95\% interval: $[0.102,0.158]$), while decreasing
the alternative's by $0.028$ nats. This is consistent with
Figure~\ref{fig:mechanism_analysis}(c), where its negation-induced
output difference suppresses variants of the original answer, ``Red.''
The same difference promotes black-related tokens rather than the
generated alternative, ``Blue,'' suggesting that suppressing the
original answer and promoting the eventual alternative need not be
performed by the same head.

Jointly restoring the complete outputs of the five heads produces
a normalized log-probability gap shift of $21.4$\%, compared with
$2.1$\% for five layer-count-matched random sets on average
(Figure~\ref{fig:mechanism_analysis}(b)).
Table~\ref{tab:app-component-answer-scores} separates the effects
on the original and alternative answers. Unlike the L31H13
answer-token intervention, joint restoration includes contributions
from all input tokens.

\begin{table}[t]
\centering
\small
\setlength{\tabcolsep}{5pt}
\begin{tabular}{@{}lrrr@{}}
\toprule
Intervention & $\Delta s(a)$ & $\Delta s(b)$ & \makecell{Normalized gap\\shift (\%)} \\
\midrule
Restore L31H13 answer-token contribution & $+0.129$ & $-0.028$ & $0.8$ \\
Restore five head outputs & $+3.438$ & $-2.336$ & $21.4$ \\
Restore 256 neuron activations & $+4.053$ & $-2.467$ & $24.7$ \\
Restore both sets & $+4.937$ & $-4.548$ & $35.2$ \\
Double 256 neuron differences & $-8.259$ & $-1.057$ & $-28.2$ \\
\bottomrule
\end{tabular}
\caption{Component interventions in Gemma 3-12B-IT, averaged across
six benchmarks. $\Delta s(a)$ and $\Delta s(b)$ are changes in the
original and alternative answers' log-probabilities (nats);
gap shift is the normalized log-probability gap shift toward the
original prompt.}
\label{tab:app-component-answer-scores}
\end{table}

\paragraph{Replication in Gemma 3-4B-IT.}
The corresponding six-benchmark experiment on 925 validation pairs
also finds an effect beyond matched random controls: restoring five
heads produces a $29.5$\% normalized log-probability gap shift, compared
with $12.0$\% for random heads (Table~\ref{tab:app-model-components}).
Head indices are selected separately for each model; the Gemma 12B
indices are not transferred to Gemma 4B. This supports a contribution
of the selected heads to answer preference in both models, without
assuming that every head has the same functional role.

\begin{table}[t]
\centering
\small
\setlength{\tabcolsep}{4pt}
\renewcommand{\arraystretch}{1.12}
\begin{tabular}{llccr}
\toprule
Model & Restored components & Match (\%) & Gap shift (\%) & \makecell{Random gap\\shift (\%)} \\
\midrule
Gemma 3-4B-IT & Five heads & 52.4 [46.1, 58.7] & 29.5 [26.2, 32.8] & 12.0 \\
Gemma 3-4B-IT & 256 neurons & 41.8 [36.6, 47.8] & 22.0 [19.2, 24.9] & -0.3 \\
Gemma 3-4B-IT & Both & 59.6 [52.8, 66.3] & 36.5 [33.3, 39.8] & 12.0 \\
Gemma 3-12B-IT & Five heads & 37.4 [32.8, 43.0] & 21.4 [18.5, 24.4] & 2.1 \\
Gemma 3-12B-IT & 256 neurons & 43.6 [39.0, 48.5] & 24.7 [22.4, 26.8] & 0.1 \\
Gemma 3-12B-IT & Both & 56.3 [50.7, 62.0] & 35.2 [32.5, 38.0] & 2.2 \\
\bottomrule
\end{tabular}
\caption{Prompt-only restoration of non-negated component values in negated prompts.  Metrics and intervals follow Table~\ref{tab:app-model-signal}. }
\label{tab:app-model-components}
\end{table}

\clearpage
\subsection{Analysis on the Answer Category Heads}
\label{app:additional-results-components_2}

We examine whether these heads support the answer category rather than a particular alternative. Across 36 validation examples from 13 relations in Gemma 3-12B-IT, L25H8's logit-lens outputs supports same-relation answer candidates both without and with negation: the mean candidate-pool uplift is $0.081$ and $0.069$, respectively. However, its output difference does not reliably identify the model's chosen alternative among these candidates. L31H12 shows a related pattern: its output difference increases the candidate-pool score by $0.180$ on average while reducing preference for the original answer. The L26H3 example in Figure~\ref{fig:mechanism_analysis}(c) similarly assigns high scores to category-related tokens such as ``color,'' ``colour,'' and ``pigment.''

We then test whether these heads affect the prediction. Removing L25H8's output on 1,540 validation pairs increases the original answer's full-sequence log-probability by $1.468$ nats and decreases the alternative's by $0.489$ nats. In a separate experiment on the validation set, restoring L31H12 and L26H3 to their non-negated outputs at the final prompt token reduces the alternative-minus-original first-token logit gap by $0.378$ and $0.125$, respectively, with 95\% bootstrap intervals of $[0.125, 0.682]$ and $[0.047, 0.208]$.    



\clearpage

\subsection{Analysis on the Negation-specific Neurons}
\label{app:additional-results-components_3}

Negation-related attribution is more concentrated in neurons than
attention changes are in heads. Among 353,280 downstream neurons,
the top 1\% (3,533 neurons) accounts for $39.7$\% of the positive
attribution score, and the top 10\% accounts for $75.5$\%
(Figure~\ref{fig:mechanism_analysis}(a)). The corresponding fractions
of total head score are $4.1$\% and $24.6$\%.

Restoring the 256 selected neurons increases the original answer's
log-probability by $4.05$ nats and decreases the alternative's by
$2.47$ nats (Table~\ref{tab:app-component-answer-scores}).
The resulting normalized log-probability gap shift is $24.7$\%,
compared with $0.1$\% for matched random neurons.
Doubling their natural activation differences instead increases the
alternative-over-original gap by $7.20$ nats. This is mainly an effect
of original-answer suppression: its log-probability decreases by
$8.26$ nats, while the alternative's also decreases, by $1.06$ nats.
Thus, amplification strengthens relative preference for the alternative
without necessarily increasing its absolute probability.
The Blue-promoting and Red-suppressing readout in
Figure~\ref{fig:mechanism_analysis}(c) illustrates this distinction
for an example.

The selected neurons are shared only partly across relations.
For the 72 benchmark--relation groups containing at least eight training
pairs, the relation-specific top 256 neurons overlap with the global
top 256 by $21.5$\% on average. 
Joint head and neuron restoration produces a $35.2$\% normalized
log-probability gap shift, larger than either intervention alone but
smaller than their sum. 

In a separate follow-up on 612 successful PopQA and RippleEdits
validation pairs, we remove the L24 negation signal at $\beta=1$ and
restore the selected components to their natural negated values.
Joint restoration reduces original-answer matching from $56.6$\%
to $17.4$\% and normalized log-probability gap shift from $32.3$\%
to $9.2$\%. Its match rate is $39.3$ percentage points lower than
matched random restoration (95\% interval: $[34.1,44.5]$). 

\paragraph{Replication and signal mediation across models.}
In Gemma 3-4B-IT, restoring 256 neurons produces a $22.0$\% normalized
log-probability gap shift, compared with $-0.3$\% for matched random
neurons. Joint head and neuron restoration increases the shift to
$36.5$\% (Table~\ref{tab:app-model-components}). Like the Gemma 12B
results, these are prompt-only interventions over six benchmarks.

\begin{figure}[p]
  \centering
  \includegraphics[width=0.9\textwidth]{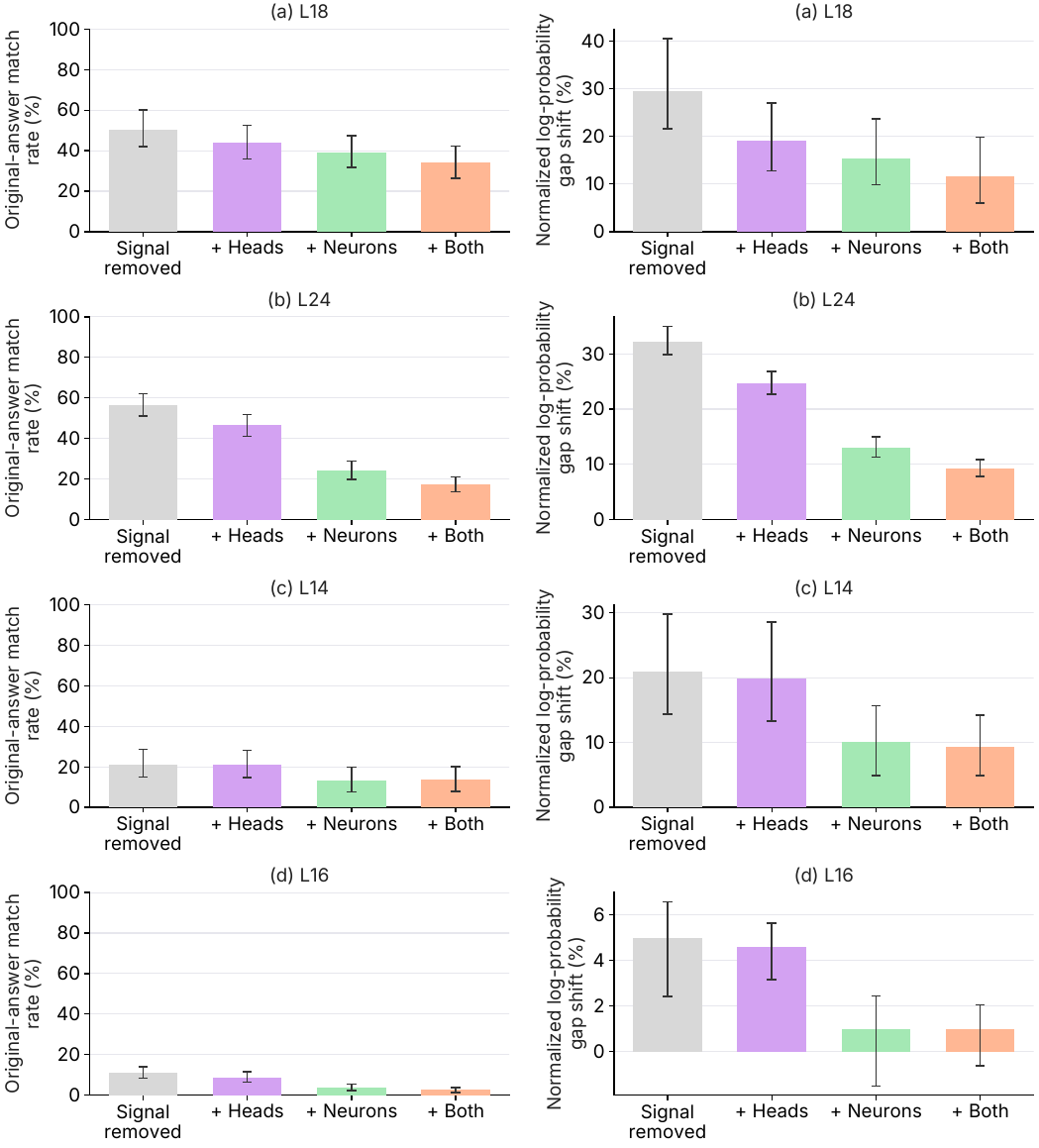}
  \caption{Signal removal followed by restoration of natural negated
  component values ($\beta=1$). Rows show (a) Gemma 3-4B-IT,
  (b) Gemma 3-12B-IT, (c) Llama 3.1-8B-Instruct, and (d) Qwen3.5-9B.
}
  \label{fig:app-model-component-mediation}
\end{figure}

We also evaluate the signal-removal and component-restoration follow-up
in all four models (Figure~\ref{fig:app-model-component-mediation};
Table~\ref{tab:app-model-mediation}). Restoring the selected heads and
neurons to their natural negated values reduces original-answer matching
relative to matched random restoration in every model. 

\begin{table}[t]
\centering
\small
\setlength{\tabcolsep}{4pt}
\renewcommand{\arraystretch}{1.12}
\begin{tabular}{lrrccc}
\toprule
Model & Layer & Pairs & \makecell{Match (\%)\\Removal $\to$ restore} & \makecell{Gap shift (\%)\\Removal $\to$ restore} & \makecell{Match difference (pp)\\Selected $-$ random} \\
\midrule
Gemma 3-4B-IT & L18 & 290 & 50.4$\to$34.3 & 29.4$\to$11.6 & -12.3 [-19.3, -6.8] \\
Gemma 3-12B-IT & L24 & 612 & 56.6$\to$17.4 & 32.3$\to$9.2 & -39.3 [-44.5, -34.1] \\
Llama 3.1-8B-Instruct & L14 & 368 & 21.3$\to$13.6 & 21.0$\to$9.3 & -7.0 [-11.3, -3.4] \\
Qwen3.5-9B & L16 & 642 & 11.1$\to$2.6 & 5.0$\to$1.0 & -7.7 [-10.0, -5.4] \\
\bottomrule
\end{tabular}
\caption{Signal removal followed by joint restoration of selected heads and neurons at their natural negated values ($\beta=1$). }
\label{tab:app-model-mediation}
\end{table}

These follow-ups use the two factual benchmarks, with continuous
interventions during generation. The frozen signal layers are L18,
L24, L14, and L16 for Gemma 3-4B-IT, Gemma 3-12B-IT,
Llama 3.1-8B-Instruct, and Qwen3.5-9B, respectively; they are not
uniformly the maxima of the prompt-only sweep.
Gemma component discovery uses six benchmarks, whereas this follow-up's
Llama and Qwen component discovery uses PopQA and RippleEdits.

\clearpage
\subsection{Extended Results on the Subspace Experiment}
\label{app:subspace_results}

Tables~\ref{tab:app-model-das-candidate}
and~\ref{tab:app-model-das-category} report the original frozen rank-4
candidate-preference and answer category spaces for all four models.
The PopQA evaluation includes non-negated prompts without filtering
by whether the generated original answer is correct. Layers and
checkpoints are fixed before validation; later layer reselection or
higher-rank runs are not substituted into these comparisons.
Unlike the prompt-only training intervention, the reported validation
replacements continue during generation, with the reference execution
conditioned on the same generated prefix.

\paragraph{Candidate preference.}
We replace subspace coordinates using another prompt from the same
category with a different answer. All four models show larger
normalized log-probability gap shifts toward that answer with the
learned space than with rank-matched random spaces.
The effect varies by model and category: Gemma's shifts range from
$29.9$\% to $38.7$\%, Llama's from $9.1$\% to $13.6$\%, and Qwen's
from $12.4$\% to $45.1$\%.
Full-residual replacement is a same-layer comparison, not an upper
bound; replacing more coordinates can also alter other useful information.

\begin{table}[t]
\centering
\small
\setlength{\tabcolsep}{4pt}
\renewcommand{\arraystretch}{1.12}
\begin{tabular}{llrrrrr}
\toprule
Model & Category & Layer & Pairs & Learned & Random & Full \\
\midrule
Gemma 3-4B-IT & City & L23 & 64 & 29.9 & 0.1 & 35.0 \\
Gemma 3-4B-IT & Person Name & L27 & 583 & 30.1 & 0.2 & 91.9 \\
Gemma 3-4B-IT & Religion & L26 & 33 & 36.9 & 0.1 & 87.9 \\
Gemma 3-4B-IT & Sport & L31 & 40 & 38.7 & 0.2 & 91.1 \\
Gemma 3-12B-IT & City & L38 & 64 & 32.2 & 0.1 & 69.8 \\
Gemma 3-12B-IT & Person Name & L42 & 583 & 32.4 & 0.2 & 74.2 \\
Gemma 3-12B-IT & Religion & L44 & 33 & 37.5 & 0.2 & 85.8 \\
Gemma 3-12B-IT & Sport & L40 & 40 & 33.9 & 0.0 & 62.4 \\
Llama 3.1-8B-Instruct & City & L13 & 64 & 13.5 & -0.0 & 13.5 \\
Llama 3.1-8B-Instruct & Person Name & L15 & 583 & 13.6 & 0.0 & 38.8 \\
Llama 3.1-8B-Instruct & Religion & L14 & 33 & 9.1 & 0.0 & 17.6 \\
Llama 3.1-8B-Instruct & Sport & L13 & 40 & 13.2 & -0.0 & 9.2 \\
Qwen3.5-9B & City & L19 & 64 & 12.4 & 0.0 & 13.0 \\
Qwen3.5-9B & Person Name & L31 & 583 & 13.9 & 0.2 & 100.0 \\
Qwen3.5-9B & Religion & L30 & 33 & 45.1 & 0.4 & 76.7 \\
Qwen3.5-9B & Sport & L24 & 40 & 31.1 & 0.3 & 62.3 \\
\bottomrule
\end{tabular}
\caption{Rank-4 candidate-space interventions between non-negated PopQA prompts with different answers in the same category. }
\label{tab:app-model-das-candidate}
\end{table}

\paragraph{Answer category.}
We replace category-space coordinates between prompts from different
categories and score each category using a fixed training answer bank.
Learned spaces produce larger category-score shifts than the random
controls (Table~\ref{tab:app-model-das-category}).
The separate within-category test produces smaller answer-identity
shifts, ranging from $1.0$\% to $4.6$\%.
These are different metrics with different denominators, not quantities
whose ratio measures how well the two kinds of information are separated.
In particular, normalized category-gap shifts can exceed $100$\%
and have wide intervals; they are not category-classification accuracies.

\begin{table}[t]
\centering
\small
\setlength{\tabcolsep}{4pt}
\renewcommand{\arraystretch}{1.12}
\begin{tabular}{lrrcrrr}
\toprule
Model & Layer & Pairs & Learned & Random & Full & \makecell{Within-category\\answer gap} \\
\midrule
Gemma 3-4B-IT & L19 & 720 & 80.1 [69.9, 89.1] & 0.2 & 70.3 & 4.2 \\
Gemma 3-12B-IT & L27 & 720 & 74.3 [32.5, 104.2] & -1.7 & 108.8 & 4.6 \\
Llama 3.1-8B-Instruct & L15 & 720 & 141.6 [58.0, 280.0] & 0.4 & 62.2 & 3.8 \\
Qwen3.5-9B & L19 & 720 & 103.5 [96.7, 111.6] & 0.0 & 67.9 & 1.0 \\
\bottomrule
\end{tabular}
\caption{Rank-4 answer-category-space interventions between non-negated PopQA prompts. }
\label{tab:app-model-das-category}
\end{table}

Results weight original questions equally within each evaluated category
and then weight categories equally. Random controls average five spaces
of the same rank. The experiments establish causal effects of the
learned coordinates under the tested replacements.

\clearpage

\subsection{Failure Diagnosis}
\label{app:dianosis}

\paragraph{Natural attention changes.}
We first compare successful and repeated-answer cases without intervening
in the model. The cached Gemma 3-12B-IT analysis contains 1,272
classified validation pairs from 400 distinct original questions in
PopQA and RippleEdits. We average forms within each question and outcome
group, then weight questions equally. A question with both outcomes
contributes to both groups; 5,000 question-bootstrap resamples preserve
this dependence.

L31H13 reduces attention to answer tokens in both groups, but the
reduction is smaller in failures: $36.3$\% to $8.6$\% in successes
and $38.1$\% to $17.2$\% in failures.
The failure--success difference after negation is $8.7$ percentage
points (95\% interval: $[7.0,10.4]$), whereas the original-prompt
difference is $1.8$ points ($[-0.4,4.0]$).
Within the 150 questions that have both outcomes across forms,
negated answer-token attention remains higher in failing forms
($16.2$\% versus $8.5$\%). 

\paragraph{Strengthening the identified changes.}
We next evaluate interventions on 606 validation pairs for which a
fresh unmodified execution reproduces the stored correct original
answer and repeats it under negation. This is a separate cohort from
the cached attention analysis, with $100$\% original-answer matching
by construction. We remove the answer-token contributions of five
heads selected by their largest training-set attention decreases,
double the 256 selected neurons' negation-induced differences, or
apply both. The suppression-head set is L31H13, L27H5, L26H13,
L27H7, and L31H6; it differs from the general attention-change ranking
used in Figure~\ref{fig:mechanism_analysis}.
Interventions continue during generation, and reference executions
use the same generated prefix. Results equally weight PopQA and
RippleEdits.

These interventions reduce original-answer matching to $94.9$\%,
$58.6$\%, and $57.6$\%, respectively.
The joint intervention's match rate is $40.9$ percentage points lower
than that of layer-count-matched random controls (95\% interval:
$[36.5,45.7]$). Thus, strengthening the identified changes can overturn
some repeated answers.

\paragraph{Relation to original-answer support.}
Finally, we ask whether these answer-token paths provide stronger
support for the original answer in failures. We measure support as
the original answer's full-sequence log-probability before minus after
removing the paths from the non-negated execution. Across the full
1,312-pair factual validation cohort, mean support is $0.084$ nats
in successes and $0.047$ nats in failures. The failure--success
difference is $-0.038$ nats (95\% interval: $[-0.124,0.053]$).

\clearpage

\begin{figure}[p]
  \centering
  \includegraphics[width=0.9\textwidth]{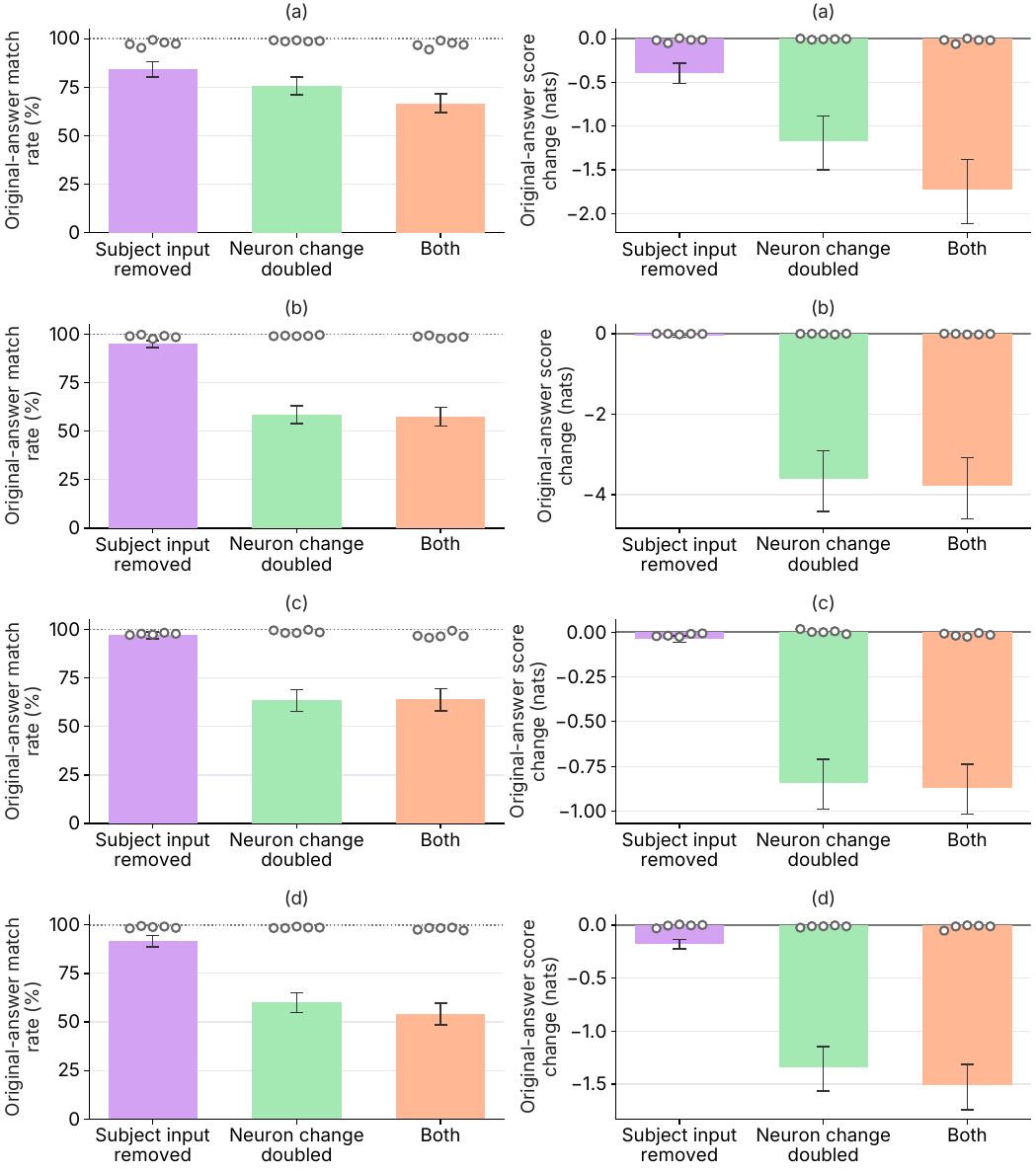}
  \caption{Interventions on repeated-answer failures in (a) Gemma 3-4B-IT,
  (b) Gemma 3-12B-IT, (c) Llama 3.1-8B-Instruct, and (d) Qwen3.5-9B.
  Left: original-answer match rate. Right: change in the original answer's
  full-sequence log-probability relative to its unmodified negated baseline.
  Interventions continue during generation; error bars show 95\%
  question-bootstrap intervals.}
  \label{fig:app-model-failure-interventions}
\end{figure}

\paragraph{Other models.}
The same intervention design reduces repetition in the other models
(Figure~\ref{fig:app-model-failure-interventions};
Table~\ref{tab:app-model-failures}). Joint intervention yields
original-answer match rates of $66.7$\% in Gemma 3-4B-IT,
$64.0$\% in Llama 3.1-8B-Instruct, and $54.3$\% in Qwen3.5-9B.
Each model is evaluated on its own current-baseline repetition cohort,
with both factual benchmarks equally weighted and interventions repeated
during generation. The joint effect exceeds that of matched random
controls in all four models. These results extend the causal finding
beyond Gemma 12B model.

\begin{table}[t]
\centering
\small
\setlength{\tabcolsep}{4pt}
\renewcommand{\arraystretch}{1.12}
\begin{tabular}{llrrrc}
\toprule
Model & \makecell{Pairs/\\questions} & Heads & Neurons & Both & \makecell{Both $-$ random\\(pp)} \\
\midrule
Gemma 3-4B-IT & 604/216 & 84.2 & 75.8 & 66.7 & -30.2 [-34.9, -25.5] \\
Gemma 3-12B-IT & 606/268 & 94.9 & 58.6 & 57.6 & -40.9 [-45.7, -36.5] \\
Llama 3.1-8B-Instruct & 473/202 & 97.2 & 63.7 & 64.0 & -33.1 [-38.8, -27.7] \\
Qwen3.5-9B & 527/206 & 91.7 & 60.0 & 54.3 & -43.7 [-49.4, -38.4] \\
\bottomrule
\end{tabular}
\caption{Original-answer match rate (\%) on current-baseline repetition failures. }
\label{tab:app-model-failures}
\end{table}

Finally, the original-prompt path-support test does not establish stronger
retrieval in failures for any of the four models
(Table~\ref{tab:app-model-path-support}). The failure--success difference
is negative in Gemma 3-4B-IT, while its interval includes zero in the
other three models. Thus, the confidence association and the causal
ability to change failed predictions remain distinct from a demonstrated
causal explanation in terms of stronger original-answer retrieval.

\begin{table}[t]
\centering
\small
\setlength{\tabcolsep}{4pt}
\renewcommand{\arraystretch}{1.12}
\begin{tabular}{lrrrc}
\toprule
Model & Pairs & Success & Failure & Failure $-$ success \\
\midrule
Gemma 3-4B-IT & 966 & 1.187 & 0.276 & -0.910 [-1.675, -0.241] \\
Gemma 3-12B-IT & 1312 & 0.084 & 0.047 & -0.038 [-0.124, 0.053] \\
Llama 3.1-8B-Instruct & 1381 & 0.079 & 0.028 & -0.051 [-0.120, 0.015] \\
Qwen3.5-9B & 1206 & 0.148 & 0.085 & -0.064 [-0.189, 0.031] \\
\bottomrule
\end{tabular}
\caption{Support for the original answer supplied by the selected answer-token paths in prompts without negation.}
\label{tab:app-model-path-support}
\end{table}

\clearpage






\subsection{Training Results}

For general capability, we report MMLU-Pro 5-shot exact match,
HellaSwag 10-shot normalized accuracy, GSM8K 8-shot CoT exact match,
and IFEval 0-shot prompt-level strict accuracy.

\begin{table}[t]
\centering
\begingroup
\small
\setlength{\tabcolsep}{5pt}
\renewcommand{\arraystretch}{1.13}
\caption{general-capability results for
Gemma-3-12B-IT checkpoints (\%).}
\label{tab:gemma12b-full-general}
\resizebox{\linewidth}{!}{%
\begin{tabular}{@{}l rrrr @{\hspace{10pt}}r@{}}
\toprule
Method & MMLU-Pro & HellaSwag & GSM8K & IFEval & General Avg. \\
\midrule
Base & 58.25 & 83.78 & 89.92 & 80.22 & 78.04 \\
\addlinespace[3pt]
\textbf{ALiT (ours)} & \textbf{57.65} & \textbf{83.44} & \underline{89.23} & \underline{80.41} & \underline{77.68} \\
\midrule
Unlikelihood & \underline{57.46} & 83.23 & \textbf{89.61} & \textbf{81.52} & \textbf{77.95} \\
SFT & 56.06 & 82.95 & 88.86 & 80.04 & 76.98 \\
DPO & 56.53 & \underline{83.42} & 89.08 & \underline{80.41} & 77.36 \\
\bottomrule
\end{tabular}%
}
\endgroup
\end{table}

\label{app:training}
\begin{figure}[t]
  \centering
  \includegraphics[width=\linewidth]{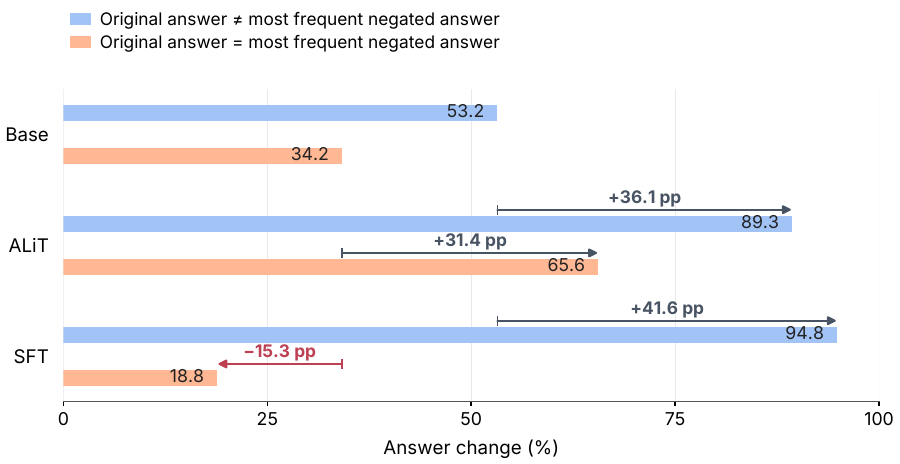}
  \caption{Answer change for Base, ALiT, and SFT, grouped by whether the original answer matches the most frequent negated answer. Arrows show percentage-point changes from Base within each group. }
  \label{fig:training-answer-bias}
\end{figure}

\clearpage

\section{Details on the Training Algorithm}
\label{app:training}

\subsection{Baselines}
\label{app:training-baselines}

We compare ALiT with supervised fine-tuning (SFT), unlikelihood
training (UL), and Direct Preference Optimization (DPO).

For a negated prompt $x_i^{\mathrm{neg}}$, let $y_i^+$ denote
the alternative used as a positive training target and $y_i^-$
the original answer used as a negative target.
For an answer sequence $y$, its conditional log-probability is
\begin{equation}
\log p_\theta(y \mid x)
=
\sum_{t=1}^{|y|}
\log p_\theta(y_t \mid x,y_{<t}).
\label{eq:baseline-sequence-probability}
\end{equation}


\paragraph{Supervised fine-tuning.}
SFT increases the probability of a specified alternative under
the negated prompt:
\begin{equation}
\mathcal L_{\mathrm{SFT}}
=
-\mathbb E_i
\left[
\log p_\theta(y_i^+ \mid x_i^{\mathrm{neg}})
\right].
\label{eq:baseline-sft}
\end{equation}
This provides direct supervision for what the model should
produce, rather than only specifying which answer to avoid.
Training with negated examples has been explored in prior work,
including ScoNe \citep{she-etal-2023-scone}.

\paragraph{Unlikelihood training.}
UL reduces the probability of the original answer under the
negated prompt.
For token-level unlikelihood, let $\mathcal T_i$ denote the
answer positions at which the penalty is applied:
\begin{equation}
\mathcal L_{\mathrm{UL}}
=
-\mathbb E_i
\left[
\frac{1}{|\mathcal T_i|}
\sum_{t\in\mathcal T_i}
\log
\left(
1-p_\theta
\left(
y_{i,t}^- \mid
x_i^{\mathrm{neg}},y_{i,<t}^-
\right)
\right)
\right].
\label{eq:baseline-ul}
\end{equation}


\paragraph{Direct Preference Optimization.}
DPO increases preference for the alternative relative to the
original answer, measured against a frozen reference model
$p_{\mathrm{ref}}$ \citep{rafailov2023direct}.
Define
\begin{equation}
r_\theta(x,y)
=
\log p_\theta(y\mid x)
-
\log p_{\mathrm{ref}}(y\mid x).
\end{equation}
The objective is
\begin{equation}
\mathcal L_{\mathrm{DPO}}
=
-\mathbb E_i
\left[
\log \sigma
\left(
\gamma
\left[
r_\theta(x_i^{\mathrm{neg}},y_i^+)
-
r_\theta(x_i^{\mathrm{neg}},y_i^-)
\right]
\right)
\right],
\label{eq:baseline-dpo}
\end{equation}
where $\sigma$ is the sigmoid function and $\gamma>0$
controls the scale of the preference objective.
DPO is a general preference-learning method rather than a
negation-specific algorithm.
Here, it tests whether learning to prefer an alternative over
the original answer provides a sufficient training signal.

\clearpage
\subsection{Training Setup}
\label{app:training-setup}

\paragraph{Training data and model.}
We post-train Gemma-3-12B-IT on the PopQA training split of our
negation benchmark. The training set contains 14,951 paired prompts
from 4,534 source questions spanning 15 relations.
Each retained negation variant constitutes a separate training pair.
All methods use the same prompt pairs for two epochs, corresponding
to 29,902 pair presentations.
Validation and test examples are not used to construct training targets.

\paragraph{Parameter-efficient optimization.}
We train LoRA adapters on the model backbone while keeping the
base parameters frozen.
Adapters are applied to the attention query, key, value, and output
projections and the MLP gate, up, and down projections.
We use rank 16, a LoRA scaling parameter of 32, and zero dropout.
Training uses AdamW with a learning rate of $10^{-4}$ and zero weight
decay. The microbatch size is four, with four gradient-accumulation
steps, giving a nominal effective batch size of 16.
The backbone is loaded in bfloat16, and gradient checkpointing is
enabled. We report the checkpoint obtained after the second epoch.

\paragraph{Preserving original predictions.}
All methods (including the three baselines) include an objective to preserve original prompt prediction.
Let $y_i^-$ denote the frozen base model's stored answer to the
original prompt $x_i$. We compute a full-vocabulary forward KL
divergence at each answer position, conditioning both models on
the same stored answer prefix:
\begin{equation}
\mathcal L_{\mathrm{orig}}
=
\mathbb E_i
\left[
\frac{1}{|y_i^-|}
\sum_{t=1}^{|y_i^-|}
D_{\mathrm{KL}}\!\left(
p_0(\cdot\mid x_i,y_{i,<t}^-)
\,\Vert\,
p_\theta(\cdot\mid x_i,y_{i,<t}^-)
\right)
\right].
\label{eq:training-original-preservation}
\end{equation}
We use a preservation coefficient of $\lambda=4$ for every method.
Losses are computed only at answer positions, using the stored answer
tokens without appending an additional end-of-sequence token.

\paragraph{ALiT targets and hyperparameters.}
For ALiT, we set the confidence-margin coefficient to $\alpha=1$
and cap the extrapolation coefficient at $\beta_{\max}=20$.
We choose the smallest coefficient satisfying the margin condition
in \S~\ref{sec:negation-training}, subject to $\beta\geq1$, and
then apply the cap.
If the natural negated prediction already satisfies the condition,
we use $\beta=1$.
If the negation-induced change is numerically zero or provides no
feasible direction for satisfying the margin, we also use $\beta=1$,
retaining the natural negated target.
Consequently, capped or infeasible targets need not satisfy the
requested margin.
The target distribution is computed over the full vocabulary at
the first answer position with temperature one.
The reverse-KL target loss has coefficient one.
The reference model remains frozen throughout training, so the
targets and confidence margins do not adapt to the student.

\paragraph{Baseline target construction.}
UL penalizes the stored original answer $y_i^-$ under the negated
prompt. SFT and DPO additionally use a donor answer from another
training question with the same relation.
For each recipient, we first sample uniformly among distinct stored
answer token sequences, excluding its own, and then uniformly among
eligible donor questions with that sequence.
The donor must come from a different source question, and donors
are resampled each epoch.
These answers are model-generated pseudo-targets rather than
human-annotated alternatives.
SFT trains on the donor answer; DPO uses it as the chosen answer
and $y_i^-$ as the rejected answer.

\paragraph{Baseline objectives and weights.}
The method-specific objectives are defined in
Appendix~\ref{app:training-baselines}.
Their coefficients are one for SFT and DPO and $0.01$ for UL,
in addition to the shared preservation term
$4\mathcal L_{\mathrm{orig}}$.
SFT and UL losses are averaged over answer tokens.
DPO instead uses summed sequence log-probabilities, with preference
scale $\gamma=0.1$.
DPO is initialized directly from the base checkpoint, rather than
an SFT checkpoint, and uses the frozen base model as its reference.

\paragraph{Evaluation.}
We evaluate in-domain performance on the PopQA test split and
out-of-domain performance by pooling the test examples from
RippleEdits, PhantomWiki, SynthWorlds, GQA, and PTR.
We report the three answer metrics defined in
\S~\ref{sec:evaluation}, together with answer diversity as defined
in \S~\ref{sec:negation-training}.
General-capability performance is the arithmetic mean of scores
on MMLU-Pro, HellaSwag, GSM8K, and IFEval.


\end{document}